\documentclass[letterpaper,twocolumn,10pt]{article}
\usepackage{usenix-2020-09}
\usepackage{fullpage}
\usepackage{amsmath,amsfonts,bm}

\def\eqref#1{equation~\ref{#1}}
\def\1{\bm{1}}

\DeclareMathAlphabet{\mathsfit}{\encodingdefault}{\sfdefault}{m}{sl}
\SetMathAlphabet{\mathsfit}{bold}{\encodingdefault}{\sfdefault}{bx}{n}

\DeclareMathOperator*{\argmax}{arg\,max}
\DeclareMathOperator*{\argmin}{arg\,min}

\usepackage[utf8]{inputenc}

\usepackage{hyperref}
\usepackage{todonotes}
\usepackage{url}

\usepackage{amsthm}
\usepackage{xfrac}

\usepackage{pifont}% http://ctan.org/pkg/pifont

\usepackage{bigdelim}

\usepackage{microtype}
\usepackage{graphicx}

\usepackage{multirow}
\usepackage{tablefootnote}
\usepackage{xspace}
\usepackage{graphicx}
\usepackage[utf8]{inputenc} % allow utf-8 input
\usepackage[T1]{fontenc}    % use 8-bit T1 fonts
\usepackage{hyperref}       % hyperlinks
\usepackage{url}            % simple URL typesetting
\usepackage{booktabs}       % professional-quality tables
\usepackage{amsfonts}       % blackboard math symbols
\usepackage{nicefrac}       % compact symbols for 1/2, etc.
\usepackage{microtype}      % microtypography
\usepackage{wrapfig}

\usepackage{mathtools}
\usepackage{amsmath,amssymb,amsfonts}
\usepackage{algorithm}
\usepackage{lipsum}
\usepackage{subcaption}

\usepackage[utf8]{inputenc}
\usepackage{pgfplots}
\DeclareUnicodeCharacter{2212}{−}
\usepgfplotslibrary{groupplots,dateplot}
\usetikzlibrary{patterns,shapes.arrows}
\tikzstyle{block} = [rectangle, draw, text width=6em, text centered, rounded corners, minimum height=4em]
\usepackage{pifont}% http://ctan.org/pkg/pifont
\usepackage{amssymb}
\usepackage[capitalise]{cleveref}

\pgfplotsset{compat=newest}
\newcommand\sample{\stackrel{\mathclap{\scriptsize\mbox{\$}}}{\gets}}

\newlength\figureheight
\newlength\figurewidth
\usepackage{cryptocode}
\usepackage{algorithm}
\usepackage{algpseudocode}
\usepackage{amsmath}
\usepackage{amssymb}
\usepackage{amsthm}
\usepackage{bm}
\usepackage{enumitem}
\setlist{topsep=1pt,parsep=0pt,partopsep=0pt}

\makeatletter
\def\munderbar#1{\underline{\sbox\tw@{$#1$}\dp\tw@\z@\box\tw@}}
\makeatother

\newcommand{\jhnote}[1]{\textcolor{blue}{}}
\newcommand{\milad}[1]{\textcolor{cyan}{}}
\newcommand{\borja}[1]{\textcolor{orange}{}}
\newcommand{\mcj}[1]{\textcolor{orange}{}}
\newcommand{\ft}[1]{\textcolor{red}{}}
\newcommand{\tas}[1]{\textcolor{green}{}}
\newcommand{\at}[1]{\textcolor{red}{}}

\newcommand{\wrt}[1]{\mathrm{d}{#1}}
\newcommand{\fn}[0]{\mathrm{FN}}
\newcommand{\fp}[0]{\mathrm{FP}}
\newcommand{\n}[0]{\mathrm{N}}

\newcommand{\fdp}[0]{$f$-DP\xspace}
\newcommand{\pld}[0]{PLD\xspace}
\newcommand{\adp}[0]{$(\varepsilon,\delta)$-DP\xspace}

\usepackage{newfloat}
\DeclareFloatingEnvironment[
    fileext=los,
    listname={List of Protocols},
    name=Protocol,
    placement=tbhp
]{protocol}

\hypersetup{
    colorlinks,
    linkcolor={red!50!black},
    citecolor={blue!50!black},
    urlcolor={blue!80!black}
}

\crefname{protocol}{Protocol}{Protocols}
\crefname{crafter}{Crafter}{Crafters}
\crefname{distinguisher}{Distinguisher}{Distinguishers}

\newfloat{crafter}{htbp}{loa}
\floatname{crafter}{Crafter} %It may be done better

\newfloat{distinguisher}{htbp}{loa}
\floatname{distinguisher}{Distinguisher} %It may be done better

\title{Tight Auditing of Differentially Private Machine Learning}
\author{
Milad Nasr$^{1}$ \quad 
Jamie Hayes$^{2}$ \quad 
Thomas Steinke$^{1}$ \quad
Borja Balle$^{2}$
\\
Florian Tramèr$^{3}$ \quad
Matthew Jagielski$^{1}$ \quad 
Nicholas Carlini$^{1}$ \quad 
Andreas Terzis$^{1}$  \\
\emph{
$^1$Google \quad 
$^2$DeepMind \quad 
$^3$ETHZ} \\
\,
\\
}\date{August 2022}
\newcounter{thm}
\newtheorem{definition}[thm]{Definition}
\newtheorem{corollary}[thm]{Corollary}

\usepackage{wrapfig}
\usepackage{mathtools}
\usepackage{amsmath,amssymb,amsfonts}
\usepackage{lipsum}

\begin{document}

\maketitle
\begin{abstract}

Auditing mechanisms for differential privacy use probabilistic means to empirically estimate the privacy level of an algorithm.
For private machine learning, existing auditing mechanisms are \emph{tight}: the empirical privacy estimate (nearly) matches the algorithm's provable privacy guarantee.
But these auditing techniques suffer from two limitations. First, they only give tight estimates under implausible worst-case assumptions (e.g., a fully adversarial dataset). Second, they require thousands or millions of training runs to produce non-trivial statistical estimates of the privacy leakage.

This work addresses both issues.
We design an improved auditing scheme that yields tight privacy estimates for \emph{natural} (not adversarially crafted) \at{describe what's natural?} datasets---if the adversary can see all model updates during training. 
Prior auditing works rely on the same assumption, which is permitted under the standard differential privacy threat model. This threat model is also applicable, e.g., in federated learning settings.
Moreover, our auditing scheme requires only \emph{two} training runs (instead of thousands) to produce tight privacy estimates, by adapting recent advances in tight composition theorems for differential privacy.
We demonstrate the utility of our improved auditing schemes by surfacing implementation bugs in private machine learning code that eluded prior auditing techniques. 

\end{abstract}
\section{Introduction}

% Machine learning models are now pervasive in many sectors that process sensitive data.

Training ML models with stochastic gradient descent (SGD) is not a privacy-preserving function.
There is ample evidence that private information from training data can be inferred by observing model parameters trained with SGD or other optimizers~\cite{carlini2019secret, carlini2022membership, yeom2018privacy, shokri2017membership, balle2022reconstructing, nasr2019comprehensive}.
There is also substantial evidence that this privacy risk increases with the number of model parameters~\cite{carlini2022quantifying, jagielski2022measuring}, a worrying fact given we are now firmly in the age of large models with hundreds of billions of parameters.

Fortunately, we can train models with differential privacy (DP) guarantees~\cite{dwork2006calibrating, abadi2016deep}, which provably upper bounds any privacy leakage of the training data.
Private training typically uses a variant of SGD referred to as Differentially Private Stochastic Gradient Descent (DP-SGD).
%tracking the privacy budget through successive updates in
DP-SGD's analysis has been conjectured to be overly conservative, 
and to provide a provable guarantee on privacy leakage that overestimates the leakage in practice~\cite{jayaraman2019evaluating}.
Nasr et al.~\cite{nasr2021adversary} partially refuted this conjecture by showing that DP-SGD's analysis gives a tight estimate of the empirical privacy leakage in some worst-case regimes (that fall under the DP threat model).
However, their tightness result only holds in a narrow and very strong adversarial model, where the adversary chooses the entire training dataset.
This leads to a natural follow-up question: 
\begingroup
\addtolength\leftmargini{-10pt}
\begin{quote}
\emph{Q1: Is DP-SGD's privacy analysis only tight for worst-case datasets?} \at{same comment about defining what natural means, especially since we say "natural follow-up question" above} \tas{The results here are, in a sense, dataset-agnostic, i.e., the gradient canary results work with \emph{any} dataset. Perhaps this is how it should be framed?}
\end{quote}
\endgroup
A further limitation of the approach of Nasr et al.---and other 
techniques for \emph{auditing} DP-SGD~\cite{jagielski2020auditing, lu2022general, zanella2022bayesian}---is the computational overhead.
Differential privacy is a probabilistic guarantee, and so empirically estimating an algorithm's privacy requires computing tight probability estimates of certain events.
Existing auditing techniques do this by running the training algorithm thousands of times---which is prohibitively expensive for large models that can cost millions of dollars to train even \emph{once}.
Our second question is thus:
\begingroup
\addtolength\leftmargini{-10pt}
\begin{quote}
\emph{Q2: Can DP-SGD's privacy leakage be tightly estimated with a small number of training runs?}
\end{quote}
\endgroup

In this work, we design a new auditing scheme for DP-SGD that resolves Q1 and Q2. Our scheme provides much tighter empirical privacy estimates compared to prior work~\cite{jagielski2020auditing ,nasr2021adversary, zanella2022bayesian, lu2022general}, which match the provable privacy leakage obtained from DP-SGD's analysis even for non-adversarially-chosen training datasets.
We further design methods to reduce the number of models that need to be trained for auditing, from tens of thousands to just \textbf{two}. Despite this massive reduction in computational overhead, our empirical privacy estimates remain tight.

Our improvements over prior auditing approaches stem from a fairly simple insight. We observe that existing auditing techniques are \emph{universal}: they make no assumption about the privacy mechanism. We show that ``opening the black box'' and tailoring our scheme to the specific privacy mechanisms used in DP-SGD results in much tighter empirical privacy estimates and with significantly fewer observations (i.e., training runs).
Intuitively, \adp---the standard formulation of DP used in DP-SGD analysis---is concerned with outcomes that have probability $O(\delta)$ and, thus, we need many training runs to observe such outcomes even once. However, we can make inferences about these rare outcomes from common outcomes by leveraging our knowledge about the privacy mechanism. As an analogy, if we know that data follows a Gaussian distribution, then we could estimate the mean and variance from a few samples, and make inferences about the tails of the distribution without ever observing those tails. \tas{I added some explanatory text here. PTAL.} \milad{lgtm}
At a technical level, we adapt existing DP auditing techniques to Gaussian DP and functional DP~\cite{dong2019gaussian} which provide a more fine-grained characterisation of the privacy leakage for specific mechanisms.  
\ft{is my above characterisation correct?}\jhnote{lgtm}
We also verify experimentally that our results agree with existing auditing techniques~\cite{nasr2021adversary, zanella2022bayesian} after sufficiently many training runs.

Our improved auditing scheme enables new applications. First, our tight characterization of the empirical privacy leakage of DP-SGD unlocks the ability to directly inspect the impact of various model and training design choices on privacy.
We explore how different choices of hyperparameters, model architecture, and the assumed attacker model impact DP-SGD's empirical privacy leakage.
%\ft{some of the above seems like a contradiction to me: when we change the model, DP-SGD's analysis doesn't change and so the provable privacy remains the same. So if our auditing is tight, we should conclude that the model selection doesn't matter for empirical privacy either.}
%\jhnote{My reading was that the claim is that this audit allows us to find the ``true'' amount of privacy leakage. The reasons we claim this is because we find that in some settings the lower bound is tight to the upper bound. We then take a leap of faith that if our auditing captures the true amount of privacy leakage, then when we inspect a new model architecture that has a smaller lower bound, it is because that model leaks less, rather than a a fault/deficiency of the audit. This might not be fair.}

Second, our tight and computationally efficient auditing enables us to (probabilistically) \emph{verify the correctness} of DP-SGD implementations.
Indeed, implementing DP-SGD is notoriously difficult: subtle privacy bugs resulting from incorrect gradient clipping or noising are common and hard to detect~\cite{DBLP:journals/corr/abs-2202-12219, githubPrngReuse, tramer2022debugging}.
Auditing can help detect such errors by showing that the implementation empirically leaks more information than it should provably allow. However, existing auditing tools are either too expensive to run~\cite{nasr2021adversary}, or provide leakage estimates that are too loose to catch the most pernicious errors~\cite{zanella2022bayesian, tramer2022debugging}.
We show that our improved auditing scheme can surface bugs that would not have been captured by prior methods~\cite{nasr2021adversary, zanella2022bayesian, lu2022general, jagielski2020auditing}.
We thus encourage developers of differentially private learning algorithms to incorporate our auditing tools into their testing pipeline.

\section{Background}

We begin with a brief background on differential privacy (DP),  private machine learning, and techniques to audit the privacy guarantees claimed under DP. %We start stating several definition used throughout this paper.

\subsection{Differential privacy}\label{ssec:dp}

Differential privacy (DP) has become the gold standard method for providing algorithmic privacy \cite{dwork2006calibrating}. 
\begin{definition}[$(\varepsilon,\delta)-$ Differential Privacy (DP)] An algorithm $\mathcal{M}$ is said to be \adp if for all sets of events $S\subseteq \text{Range}(\mathcal{M})$ and all neighboring data sets $D, D' \in \mathcal{D}^n$ (where $\mathcal{D}$ is the set of all possible data points) that differ in one sample we have the guarantee:

\begin{equation}
    \Pr[\mathcal{M}(D)\in S] \leq e^{\epsilon} \Pr[\mathcal{M}(D')\in S] + \delta
\end{equation}
\end{definition}
Informally in the context of machine learning, if a training algorithm $\mathcal{M}$ satisfies \adp then an adversary's ability to distinguish if $\mathcal{M}$ was run on $D$ or $D'$ is bounded by $e^{\epsilon}$, and $\delta$ is the probability that this upper bound fails to hold. 

\paragraph{Trade-off functions and functional DP.}
There are other useful formalisms of DP.
For example, functional differential privacy ($f$-DP)~\cite{dong2019gaussian} originates from a hypothesis testing interpretation of differential privacy~\cite{wasserman2010statistical, kairouz2015composition}, where an adversary aims to distinguish $D$ from $D'$ given the output of the privacy mechanism.
Although our results will be framed using \adp, our auditing framework will operate using functional DP.
Consider the following hypothesis testing problem, given some machine learning model $f$:
\begin{align*}
    &H_0\text{: the model $f$ is drawn from } P\\ &H_1\text{: the model $f$ is drawn from } Q
\end{align*}
where $P$ and $Q$ are the probability distributions $\mathcal{M}(D)$ and $\mathcal{M}(D')$, respectively.
If $\mathcal{M}$ is differentially private, we can derive a bound on an adversary's power $1-\beta$ (i.e. True Positive Rate or TPR, where $\beta$ is the False Negative Rate or Type II error) for this hypothesis test at a significance level $\alpha$ (i.e. False Positive Rate, FPR, or Type I error). 
For example, \adp upper-bounds the power of this hypothesis test by $e^{\epsilon}\alpha  + \delta$.

Dong et al.~\cite{dong2019gaussian} define a \emph{trade-off function} to capture the difficulty in distinguishing the two hypotheses above in terms of the adversary's type I and type II errors.
Consider a rejection rule $0\leq\phi(f)\leq 1$ that takes as input the model $f$ trained by the mechanism $\mathcal{M}$, and which outputs a probability that we should reject the null hypothesis $H_0$. This rejection rule has type I error $\alpha_{\phi}=\mathbb{E}_P[\phi]$ and type II error $\beta_{\phi}=1 - \mathbb{E}_Q[\phi]$, which gives rise to the following trade-off function:

\begin{definition}[Trade-off function~\cite{dong2019gaussian}] For any two probability distributions $P$ and $Q$ on the same space define the trade-off function $T(P,Q): [0,1] \xrightarrow[]{}[0,1]$ as 
\begin{equation}
    T(P,Q)(\alpha) = \inf\ \{\beta_{\phi} : \alpha_{\phi} \leq \alpha \}
\end{equation}
where the infimum is taken over all rejection rules $\phi$.
\end{definition}

The trade-off function completely characterizes the boundary of achievable type II errors at a given significance level $\alpha$, and the optimal test is given by the Neyman-Pearson Lemma.
For arbitrary functions $f,g$ defined on $[0, 1]$, we say that $f\geq g $ if $f(\alpha) \geq g(\alpha)$ for all $\alpha\in[0,1]$.
Then, if $T(P, Q) \geq T(\tilde{P} , \tilde{Q})$, this means the distributions $P$ and $Q$ are harder to distinguish than $\tilde{P}$ and $\tilde{Q}$ at any significance level. Thus, a privacy mechanism that produces distributions $P$ and $Q$ on neighboring datasets is strictly more private than one that produces distributions $\tilde{P}$ and $\tilde{Q}$.
Dong et al.~\cite{dong2019gaussian} introduce the following formulation of differential privacy using this insight:

\begin{definition}[$f$-differential privacy ($f$-DP)] Let $f$ be a trade-off function. A mechanism $\mathcal{M}$ is $f$-DP if 
\begin{equation}
    T(\mathcal{M}(D),\mathcal{M}(D')) \geq f
\end{equation}
for all neighboring datasets $D$ and $D'$.
\end{definition}

Dong et al.~show that \adp is equivalent to \fdp for the following trade-off function:
\begin{equation}
    f_{\varepsilon,\delta}(\alpha) = \max \{0, 1-\delta-e^{\varepsilon}\alpha,e^{-\varepsilon}(1-\delta-\alpha) \}
    \label{eq:eps_delta_tradeoff}
\end{equation}

%\paragraph{Gaussian DP.}
When the underlying distributions $P, Q$ are Gaussian, we get a special case of \fdp called Gaussian DP (GDP)~\cite{dong2019gaussian}:

\begin{definition}[$\mu$-Gaussian Differential Privacy ($\mu$-GDP)] A mechanism $\mathcal{M}$ is $\mu$-GDP if 
\begin{equation}
\hspace{-1pt}T(\mathcal{M}(D),\mathcal{M}(D'))(\alpha) \geq \Phi(\Phi^{-1}(1-\alpha)-\mu) \;, \forall \alpha \in [0,1]\hspace{-2pt}
\end{equation}
for all neighboring datasets $D$ and $D'$, where $\Phi$ is the standard normal CDF.
\end{definition}

One of the main advantages of GDP is that composition of differential privacy guarantees becomes simple, the composition of two mechanisms following $\mu_1$-GDP and $\mu_2$-GDP satisfies $\mu$-GDP with $\mu=\sqrt{\mu_1^2+\mu_2^2}$.
A final fact that will be useful throughout the paper is that it is possible to interpret $\mu$-GDP in terms of \adp:

\begin{corollary}[$\mu$-GDP to \adp conversion~\cite{dong2019gaussian}]\label{eq:gdp_dp} A mechanism is $\mu$-GDP iff it is $(\varepsilon,\delta(\epsilon))$-DP for all $\varepsilon \geq 0$, where:

\begin{equation}
    \delta(\varepsilon) = \Phi\left(-\frac{\varepsilon}{\mu}+\frac{\mu}{2}\right) - e^{\varepsilon} \Phi\left(-\frac{\varepsilon}{\mu}-\frac{\mu}{2}\right)
\end{equation}
\label{eps_gdp_convert}
\end{corollary}

% \begin{definition}[Privacy Loss Distribution~\cite{}] For any two continuous distribution $\mu_{up}$ and $\mu_{lo}$, their privacy random var
% \begin{equation}
%     \mathcal{L}_{\mu_{up}/\mu_{lo}}(o):= 
% \end{equation}
 
% \end{definition}

\subsection{Differentially Private Machine Learning}

Stochastic gradient descent (SGD) can be made differentially private through two modifications: clipping individual gradients to maximum Euclidean norm of $C$ and adding random noise to the average of a batch of gradients; this algorithm is commonly referred to as DP-SGD.
Intuitively, clipping bounds the individual contribution any sample can make to the model parameters, $\theta$, and adding random noise serves to obfuscate the contributions of any individual example.
In practice the update rule for DP-SGD is given as follows: let $B$ denote a batch of examples sampled independently from a dataset $D$, each with probability $q$, $\ell$ be a loss function, and $\eta$ a learning rate, then
\begin{equation}
    \label{eqn:dpsgd}
    \theta \gets \theta - \eta \left(\mathcal{N}(0, \sigma^2I) + \frac{1}{|B|}\sum_{z \in B}  \text{clip}_{C}\left(\nabla_{\theta}\ell(\theta, z)\right)\right)
\end{equation}
where $\text{clip}_C(v)$ projects $v$ onto the $\ell_2$ ball of
radius $C$ with
\[\text{clip}_{C}(v) = v \cdot \min\left\{1, \frac{C}{\|v\|_2}\right\}.\]

When we refer to a \emph{privatized gradient}, we mean the gradient after it has been clipped, then averaged, and then noised.
To achieve  \adp, typically $\sigma$ is typically on the order of $\Omega(q\sqrt{T\log(\sfrac{1}{\delta})}\varepsilon^{-1})$~\cite{abadi2016deep} where $T$ is the number of gradient descent iterations, but tighter bounds for a given $\sigma$ have been found~\cite{mironov2017renyi, dong2019gaussian, koskela2020computing}. 
%\shs{haven't set $\sigma$ here... do we want to set it? as the actual dpsgd analysis is done with rdp not dp.}\todo{yeah we need to at least explain how sigma is chosen}
%
Each iteration of DP-SGD satisfies a particular \adp guarantee through the subsampled Gaussian Mechanism~\cite{abadi2016deep}---a composition of data subsampling and Gaussian noise addition. Since DP is immune to post-processing, we can compose this guarantee over multiple updates to reach a final \adp guarantee.

Unfortunately, a naive composition—by summing the $\epsilon$’s from each iteration—gives values of $\epsilon\gg10^4$ for accurate neural networks.
This yields a trivial upper bound of $\approx 1$ on the true positive rate for the hypothesis testing problem discussed in~\Cref{ssec:dp}, for any reasonable value of $\alpha$. 
Such a large $\epsilon$ thus does not guarantee any meaningful privacy. 
As a result, many works have proposed more sophisticated methods for analyzing the composition of DP-SGD iterations, which can prove much tighter values of $\epsilon<10$ for \emph{the same algorithm}~\cite{abadi2016deep, dong2019gaussian,mironov2017renyi,koskela2020computing}.

\subsection{Auditing DP-SGD}

Any differentially private algorithm $\mathcal{M}$ bounds an adversary's ability to infer if $\mathcal{M}$ was trained with $D$ or $D'$.
Kairouz \emph{et al.}~\cite{kairouz2015composition} show that if $\mathcal{M}$ is \adp then it defines a \emph{privacy region} (a bound on an attacker's TPR and FPR) given by
\begin{equation}
\begin{split}
   \hspace{-8pt}\mathcal{R}(\epsilon, \delta) = \{(\alpha, \beta) \mid\ &\alpha + e^{\epsilon}\beta \geq 1-\delta \land
   e^{\epsilon} \alpha + \beta \geq 1-\delta\ \land \\
   & \alpha + e^{\epsilon}\beta \leq e^{\epsilon}+\delta \land
    e^{\epsilon}\alpha + \beta \leq e^{\epsilon}+\delta \}\hspace{-2pt}
\end{split}\label{eqn:eps_region}
\end{equation}

In other words, an \adp algorithm implies a valid region for the type I ($\alpha$) and type II ($\beta$) error of any test.

The goal of a privacy \emph{audit} is to design a hypothesis test that distinguishes $D$ from $D'$ while minimizing $\alpha$ and $\beta$. Then, we can compute the privacy budget $\epsilon$, for any fixed value of $\delta$, using \Cref{eqn:eps_region} (or as we will see later, via other means).
In practice, for many interesting differentially private algorithms including DP-SGD, one cannot compute the minimum possible values of $\alpha$ and $\beta$ in closed form, and so we must rely on empirical estimates.
This is done by designing a \emph{distinguisher} that predicts if mechanism $\mathcal{M}$ operated on $D$ or $D'$. 
We then run the distinguishing experiment multiple times (i.e., by running $\mathcal{M}$ multiple times to train a model on a random choice of $D$ or $D'$), collect these observations, and compute empirical lower and upper bounds $\alpha\in(\munderbar{\alpha}, \bar{\alpha})$ and $\beta\in(\munderbar{\beta}, \bar{\beta})$ using a binomial proportion confidence interval. 
Nasr et al.~\cite{nasr2021adversary} use the Clopper-Pearson method to find $\bar{\alpha}$ and $\bar{\beta}$, ultimately deriving an empirical lower bound to $\epsilon$ by appealing to \Cref{eqn:eps_region} and noting that
\begin{equation}
   \epsilon^{\text{lower}}_{\text{emp}} = \max\left\{\ln\left(\frac{1-\bar{\alpha}-\delta}{\bar{\beta}}\right), \ln\left(\frac{1-\bar{\beta}-\delta}{\bar{\alpha}}\right), 0\right\} \label{eq: nasr_lb}
\end{equation}

The lower bound $\epsilon^{\text{lower}}_{\text{emp}}$ comes with an empirical level of confidence through the confidence level for $\bar{\alpha}$ and $\bar{\beta}$. 
% The tighter this confidence interval is, the tight our lower bound becomes.
Unfortunately, a high level of confidence in $\epsilon^{\text{lower}}_{\text{emp}}$ often requires thousands or millions of observations.

The adversary is also free to design $D$ and $D'=D\cup\{z\}$ in any way they choose, because the privacy guarantee of DP must hold for \emph{any} pair of neighboring datasets.
The goal of the auditor/adversary is thus to design $D$ and $z$ in such a way that it is easy to design a distinguisher for $\mathcal{M}(D)$ and $\mathcal{M}(D')$.

Nasr \emph{et al.} \cite{nasr2021adversary} showed that $\epsilon^{\text{lower}}_{\text{emp}}$ is close to the upper bound $\epsilon$ output by a DP accounting mechanism when $D=\emptyset$ and so the model is trained on either zero points, or one point $z$. 
That is, they designed a test where $\alpha$ and $\beta$ are minimized under this setting. 
In summary, this auditing mechanism has shown that current DP accounting methods are nearly tight~\cite{abadi2016deep, dong2019gaussian,mironov2017renyi,koskela2020computing}, by showing that the lower bounds for $\epsilon$ one can find through a statistical test are close to the upper bound for $\epsilon$ given by DP accounting.
The drawback of this analysis is that it only demonstrates the analysis is tight with a worst-case dataset, $D=\emptyset$,
%\at{is tight with datasets that have either zero or one point?}\jhnote{changed the wording} 
and to show this it is necessary to train the model thousands of times in order to find non-trivial lower bounds $\epsilon^{\text{lower}}_{\text{emp}}$.

Zanella-Béguelin et al.~\cite{zanella2022bayesian} propose a refined Bayesian approach to finding an empirical lower bound for $\epsilon$ through a non-informative prior on $(\alpha, \beta)$.
%While Nasr et al.~\cite{nasr2021adversary} use \Cref{eq: nasr_lb} to directly convert empirically computed $(\alpha, \beta)$ into a lower bound for $\epsilon$ at a fixed $\delta$, future work 
Specifically, they define a lower bound for $\epsilon$ as
\begin{equation}
\begin{split}
   \munderbar{\epsilon} = \sup\{\epsilon\in \mathbb{R}_{>0} \mid (\alpha, \beta)\notin \mathcal{R}(\epsilon, \delta)\}
\end{split}
\end{equation}

From here, they define $f_{(\alpha, \beta)}$ to be the density function of the posterior joint distribution of $(\alpha, \beta)$ given the observed trials (found through training on $\mathcal{M}(D)$ and $\mathcal{M}(D')$ multiple times).
A 100(1-$\gamma$)\% credible interval [$\munderbar{\epsilon}$, $\bar{\epsilon}$] is then defined as
\begin{equation}
\begin{split}
   & \munderbar{\epsilon} = \argmax_{\epsilon} \int\int_{\mathcal{R}(\epsilon, \delta)} f_{(\alpha, \beta)}(x,y) \mathop{\wrt x} \mathop{\wrt y} \leq \frac{\gamma}{2} \\ 
   & \bar{\epsilon} = \argmin_{\epsilon} \int\int_{\mathcal{R}(\epsilon, \delta)} f_{(\alpha, \beta)}(x,y) \mathop{\wrt x} \mathop{\wrt y} \geq 1 - \frac{\gamma}{2}
\end{split}\label{eq: msr_bound}
\end{equation}

There are number of subtle assumptions made in this approach which require unpacking.
\Cref{eq: msr_bound} cannot be evaluated in closed form, and so we must approximate it, and it is not clear how this approximation translates into a statistically sound lower bound.
Moreover, the comparison between bounds found through this method and through Clopper-Pearson may be slightly unfair, as they are distinct statements about uncertainty of an estimate.
Nevertheless, Zanella-Béguelin et al.~\cite{zanella2022bayesian} show that their method dramatically improves the tightness of the lower bound estimate for $\epsilon$ in practice. Thus, the number of training runs needed for the audit is also significantly reduced.
% \jhnote{One of my main confusions around this method is how they justify splitting the joint posterior. My reading is they do $P(\alpha, \beta| \text{observations})=P(\alpha| \text{number of false positives})P(\beta| \text{number of false negatives})$. But I don't see a proof or an strong argument for this beyond a throwaway statement. Curious to hear your thoughts? NB multiple authors on that paper are on the PC.}

In other recent work, Lu et al. \cite{lu2022general} compute an $\epsilon$ lower bound by replacing the Clopper-Pearson method for finding bounds on $(\alpha, \beta)$ with the Katz-log confidence interval~\cite{katz1978obtaining}, which directly bounds the ratio of binomial proportions and empirically gives tighter estimates for $\epsilon^{\text{lower}}_{\text{emp}}$ with fewer observations (i.e. the method requires fewer number of models that must be trained on $D$ and $D'$).
However, the Katz-log method gives a confidence bound on the ratio of $\alpha$ to $\beta$, and it is not clear if this is valid for \adp where the ratio would change to $\frac{\alpha-\delta}{\beta}$.
Lu et al.~set $\delta=0$ in their experiments, giving a lower bound for $(\epsilon, 0)$-DP.
Lu et al.~also suggest that empirical privacy leakage is dataset dependent.
Our work directly contradicts this claim; we argue that their observations were mostly due to using weaker attacks than are permitted under the DP threat model.
By instantiating a more powerful attack our results in \Cref{sec:experiments} show that the empirical privacy leakage is close to the theoretical $\epsilon$ across a range of datasets.

We note that throughout this work we use the terms \emph{trainer}, \emph{auditor}, and \emph{attacker} interchangeably.
The party that \emph{audits} the model takes on the role of an \emph{attacker} to measure the empirical privacy leakage, and this involves \emph{training} the model.

\section{Motivation \& Threat Model}\label{sec:threat_model}

% As deep learning models are getting deployed in practice, privacy leakage of the training dataset is one of the main concern. Recent works shows that models with larger models leak more information about the training dataset and in the recent years the models are getting larger~\cite{}. One of the main theoretical frameworks to ensure the privacy of the users is differential privacy. In particular, DP-SGD~\cite{} is used to train deep learning models with differential privacy.

% Many believe that the theoretical bound given by DP-SGD are very conservative and in in practice the privacy leakage is smaller. Therefore, it is important to understand the relationship between the theoretical guarantees offered by differential privacy and the measurable practical privacy metrics. Previous works~\cite{} used auditing techniques to show the relation between theoretical privacy bounds and the empirical privacy attacks. However, except in highly adversarial cases they weren't able to achieve tight bounds.

The goal of our work is to improve the efficiency of empirical privacy estimation.  
This allows us to study the gap between theoretical and practical privacy bounds. 
And, as a practical application, auditing methods can be used to validate the correctness of a DP implementation.
\at{sentence too long, consider splitting}\tas{split into three.}

Our empirical privacy estimates depend upon the specific threat model we instantiate the test within, and assumptions we place on the adversary. 
Nasr et al.~\cite{nasr2021adversary} describe several threat models and settings for auditing machine learning with differential privacy.
We similarly study multiple threat models, as there is inevitably a trade-off between the power of the audit (with a powerful adversary) and generalizability of the audit to practical machine learning applications (where the assumptions we make to instantiate a powerful adversary may be unrealistic).
We focus on three threat models in decreasing order of attack power.

\paragraph{White-box access with gradient canaries:}  
This is the main threat model considered by the DP-SGD theoretical analysis, and matches the (implicit) threat model assumed by DP. 
The adversary has access to the privatized gradient and model parameters in every update step and can choose an arbitrary gradient at each update step~\cite{balle2022reconstructing}, which we refer to as a \emph{canary} gradient. This canary gradient then gets included into the update with probability $q$. This mimics an adversary who has access to all aspects of training other than the randomness used in noise addition and batch selection (i.e., the knowledge of when $z$ was used in training, where $D'=D\cup\{z\}$).
%\ft{this is not quite correct: the adversary gets to pick the canary gradient but that gradient is only sampled into the batch with probability $p$. And so in particular, the adversary does *not* know when $z$ is used for an update.}\milad{edited, does it still read that way? }\tas{edited}

\paragraph{White-box access with input-space canaries:} 
The threat model above assumes the adversary can choose an arbitrary \emph{gradient} that is sampled into a batch of updates. This may be an unrealistic capability for an adversary in practice. Our second threat model removes this assumption, and instead allows adversaries access to intermediate updates, but restricts them to choose an arbitrary training sample (from which gradients are subsequently computed), rather than the ability to choose a gradient directly.
We refer to the training sample chosen for each update step as the \emph{canary} sample.
This setting matches the threat model of federated learning particularly well, where an adversary can access model updates but may not always have the ability to insert arbitrary gradients into the training pipeline.
\ft{isn't FL closer to the first threat model, since users send gradients directly to the central server?}\tas{This is a bit weird: why would the adversary control the target's gradients? It seems like the adversary has already won if they have this power.}
%\ft{same as above}\tas{The first sentence makes it sound like it's realistic to assume the adversary can insert an arbitrary training example. But this is rather weird in practice. Do we want to say anything about this?}\milad{I am a bit confused, do you mean that the data should come from an underlying distribution? }

\paragraph{Black-box access:} One of the most restrictive threat models to conduct audits on is that of an adversary who can only insert a training example at the beginning of training, and observe the model after it has completed training. 
In other words, the adversary does not get to observe or influence intermediate model updates.
While this is the most restrictive setting from an adversarial perspective, it is perhaps the most realistic from a practical standpoint. 
We stress that this threat model is not the typical setting analyzed in DP, which assumes intermediate model updates are visible to the adversary. We choose to evaluate it because it allows us to compare how the incremental removal of adversarial access to model updates and gradients affects the tightness of our lower bound for $\epsilon$.

% \begin{itemize}
%     \item White-box with Gradient attack
%     \item White-box with Input attack
%     \item Black-box attack: 
% \end{itemize}
% we consider thread models that can be used to audit DP-SGD. We follow the adversary instantiating approach~\cite{} and we evaluate the effect of different settings and threat models (as mentioned in Section~\ref{sec:threat}). 

\section{Auditing with \fdp}

\borja{Would be nice to include a (meta-)algorithm in this section explaining how the different algorithms in the rest of the paper are combined to produce lower bounds.}
To audit the privacy of DP-SGD, an adversary repeatedly runs a distinguishing attack to infer if a model's training set was either $D$ or $D'$; by measuring the false positive and false negative rates of the attack we can bound privacy.
All prior work has used $(\varepsilon, \delta)$-DP definition to audit the privacy parameters of the algorithm. 
The limitation of this approach is that different differential privacy mechanisms with \emph{identical} $(\varepsilon, \delta)$ guarantees can have \emph{different} trade-offs between false positive and false negative rates, which are upper bounded by the trade-off function of $(\varepsilon, \delta)$-differential privacy.
In particular, any $(\epsilon, \delta)$-DP guarantee corresponds to two symmetric supporting linear functions defining the trade-off between type I and type II errors.
However, any mechanism will have its own ``true'' trade-off curve capturing the relationship between the FPR and TPR of the best possible attack on the mechanism, consisting of the intersection between a collection of $(\epsilon, \delta(\epsilon))$-DP curves where $(\epsilon, \delta(\epsilon))$ satisfy \Cref{eps_gdp_convert}.
% \todo{Nicholas: you haven't defined adp at all. I don't know why you're saying we don't use somthing that's not defined.\milad{are you refereing to the \adp defintion ? we had it in section 2.1, but I rewrite it to make it a bit more obvious}}
While it is possible to audit any algorithm to lower bound its privacy with an \adp guarantee, we instead use an \fdp guarantee that is as close as possible to the true trade-off function of the mechanism.
By doing so, we can avoid any looseness that appears in converting between this \fdp guarantee and its collection of many \adp guarantees.
%A $\mu$-GDP mechanism's trade-off curve is instead given by the intersection of the collection of $(\epsilon, \delta(\epsilon))$-DP curves where $(\epsilon, \delta(\epsilon))$ satisfy \Cref{eps_gdp_convert}.
%By auditing with $\mu$-GDP we can exploit the fact that this formulation of differential privacy will give tighter control over the trade-off between type I and type II errors with a single parameter $\mu$ rather than attempting to lower bound the collection of $(\epsilon, \delta(\epsilon))$-DP guarantees simultaneously.
%\borja{My perspective on this is a bit different: most mechanisms satisfy a collection of guarantees $(\epsilon(\delta), \delta)$-DP or $(\epsilon, \delta(\epsilon))$-DP for varying values of $\delta$ or $\epsilon$, and previous auditing methods fail to exploit that. For example, if you take the "polygonal" regions coming from fixed parameter DP and intersect them all for all the valid values $(\epsilon, \delta(\epsilon))$-DP you will recover the more precise mechanism-specific curves from \Cref{fig:diff_mechanism_tradeoff}. Using, eg, GDP gives you a tighter curve controlled by a single parameter, which is why it's advantegeous for auditing instead of trying to lower bound all the $(\epsilon, \delta(\epsilon))$-DP curves at once.}
%\jhnote{Added something}
%\mcj{one way to frame this could be to talk about the ``true'' tradeoff function of a mechanism, and then later mention that the true tradeoff function for our attacks is much closer to gdp than it is to eps delta}

\begin{figure}
    \centering
    \includegraphics[scale=0.35]{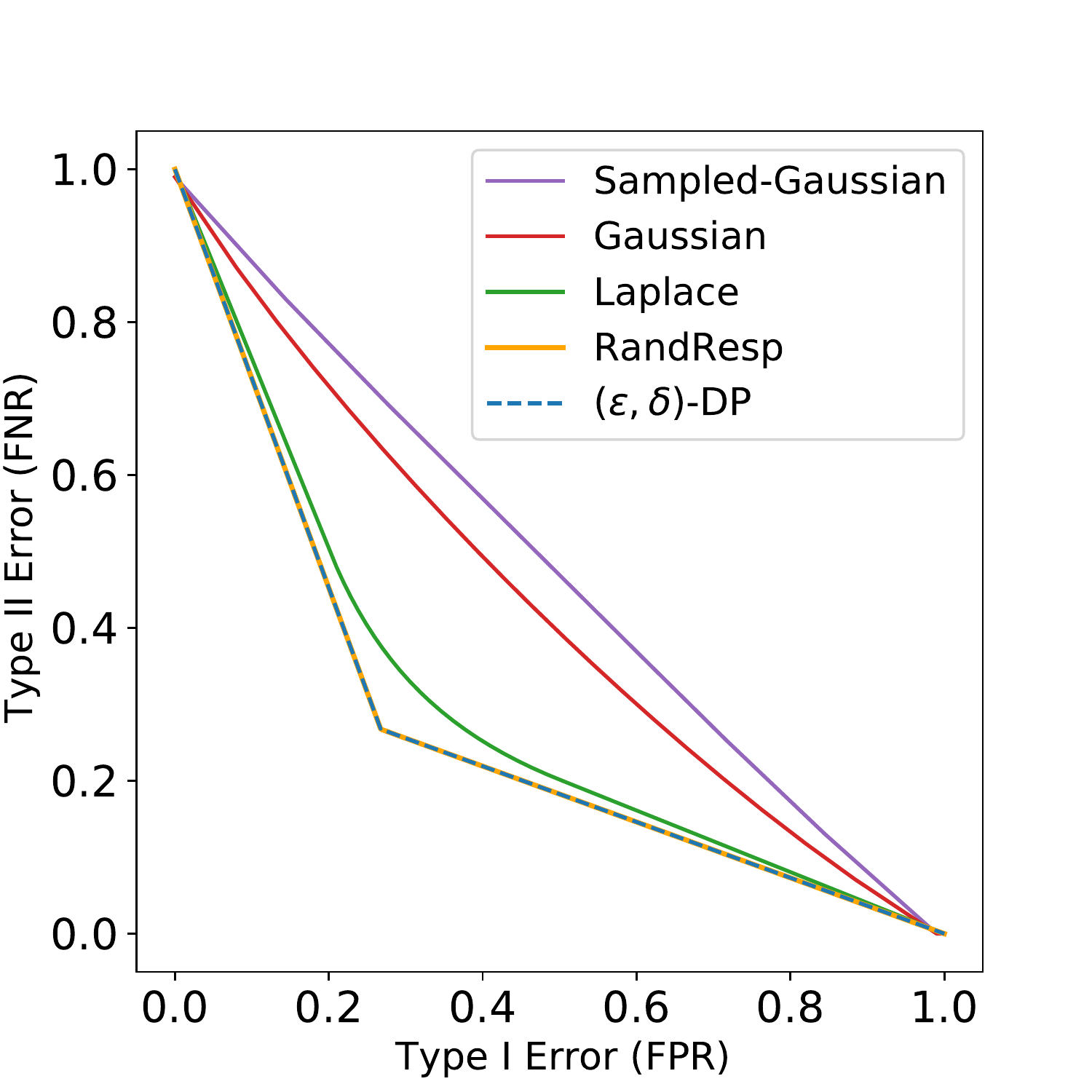}
    \caption{Comparison of the trade-of functions for different DP mechanisms that satisfy \adp where $\varepsilon=1,\delta=10^{-5}$. Note that the trade-off curve for the Random-Response mechanism overlaps with \adp.}
    \label{fig:diff_mechanism_tradeoff}
    \vspace{-0.7cm}
\end{figure}

To illustrate this idea, in \Cref{fig:diff_mechanism_tradeoff} we plot the trade-off functions for several different DP mechanisms that all satisfy \adp where $\varepsilon=1,\delta=10^{-5}$. 
Clearly, the achievable false positive and false negative rates by an adversary who wants to audit a \adp guarantee depends significantly on the underlying privacy mechanism. 

\begin{figure}[t]
    \centering
    \includegraphics[scale=0.35]{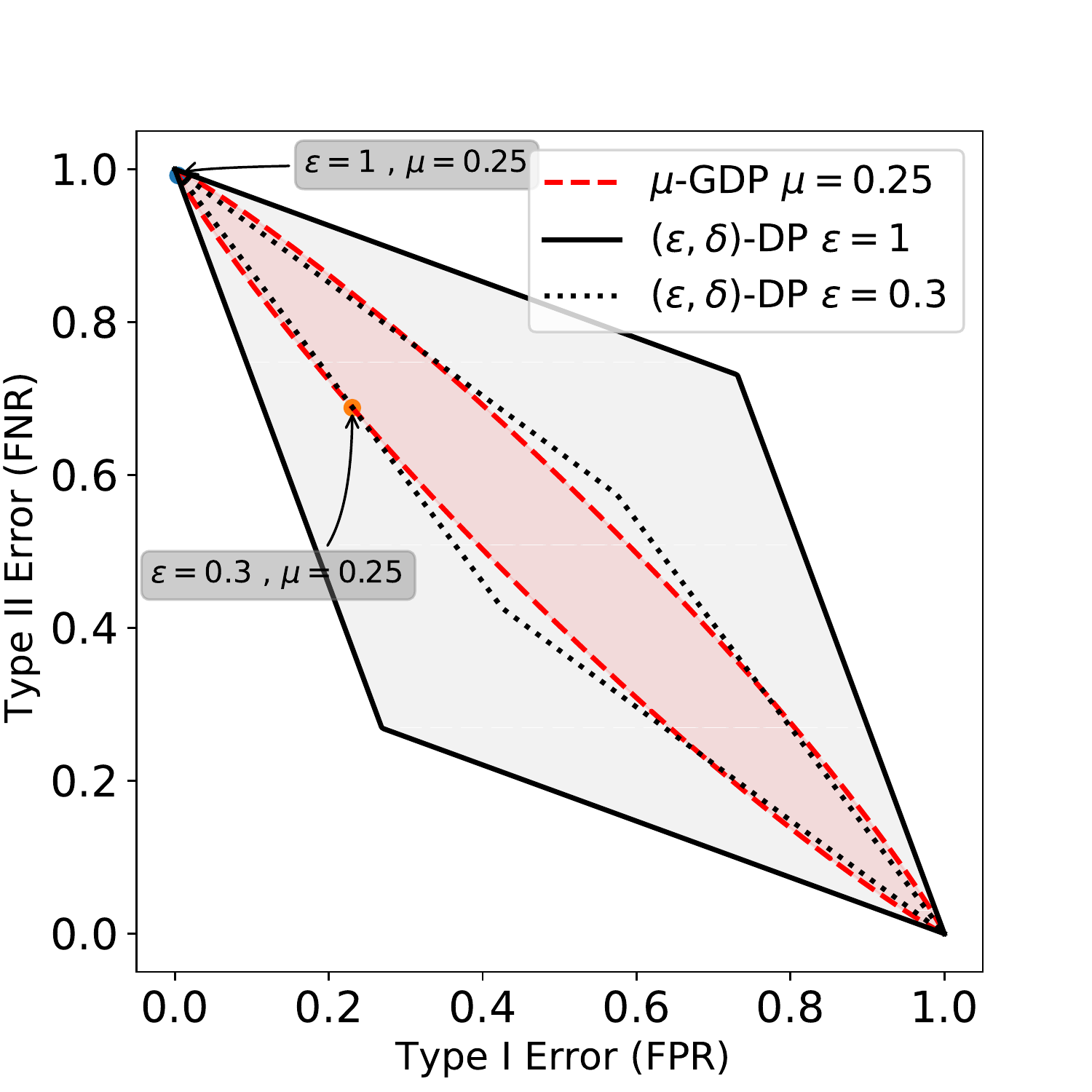}
    \caption{Comparison of the privacy region of \adp vs \fdp for the Gaussian mechanism (GDP) with the same $\varepsilon$ budget.}
    \label{fig:dp_vs_fdp}
\end{figure}

Let us now give a concrete example demonstrating the benefit of auditing by measuring the privacy region of the private mechanism directly, rather than focusing on the privacy region specified by \adp. 
Suppose we want to audit an instance of the Gaussian mechanism satisfying $(1,10^{-5})$-DP; Figure~\ref{fig:dp_vs_fdp} illustrates the privacy regions of a generic private mechanism with $\varepsilon=1,\delta=10^{-5}$ and with a Gaussian mechanism which has an equivalent \adp guarantee (i.e, $\mu$-GDP with $\mu\approx0.25$). 
% As we can see in the figure, the privacy  region of a Gaussian mechanism is different from an unspecified mechanism that satisfies the same \adp privacy guarantee.
Now, if we want to audit the Gaussian mechanism by bounding \adp, our attack needs to have very low false positive or false negative rates (corresponding to the four locations where the GDP region and the \adp region have tangent borders). 
As a concrete example, suppose we are auditing this Gaussian mechanism (with $\varepsilon=1$) and we have an attack that achieves FPR$\approx0.23$ and FNR$=1- $FPR$ (e^{0.3}+\delta)$ over an infinite number of trials (the red dot in Figure~\ref{fig:dp_vs_fdp}).
If we use \adp to audit this mechanism we get an empirical $\epsilon$ of $0.3$, and we might (incorrectly) conclude that our mechanism is not tight. 
This is not because our attack is weak, but rather it is because our attack has a large FNR and no attack can achieve a lower FNR from the definition of the Gaussian mechanism (i.e, if an attack can achieve a lower FNR it violates $\sim0.25$-GDP). 

% If one can design an attack that reaches the optimal FPR, then can use either \adp or $\mu$-GDP to audit the mechanism, this might not be possible in practice. 
% First, to have an attack we a very small FPR, we need many trials to be able to get to even calculate the small FPR. 
% Second, working with high precision have potential risk of numerical instability and error which make it difficult to work with. 
% Moreover, most of the existing works use membership inference attacks~\cite{} which are optimized for average case accuracy~\cite{}.

\subsection{Lower Bounding \fdp With Clopper-Pearson}\label{ssec:lb_with_cp}
% \todo{Nicholas: It's not obvious to me that you're still doing f-DP but with clopper pearson. I'd start by making that obvious.}
Previous works use \Cref{eq: nasr_lb} which describe the predictive power of an adversary auditing with \adp to compute the privacy parameters from the false positive and negative rates. 
However, as we have already seen by appealing to the hypothesis testing interpretation of DP, the predictive power of the adversary is by definition equal to the trade-off function of the privacy mechanism. 
Therefore, instead of using \Cref{eq: nasr_lb} to compute the privacy parameters, we can directly use the trade-off function. 
Now by upper bounding the false positive ($\alpha$) and false negative rates ($\beta$) (referred to as $\bar{\alpha},\bar{\beta}$) we can calculate the lower bound on the privacy of the mechanism. 
Similar to the previous works~\cite{nasr2021adversary} we can use the Clopper-Pearson method to compute the upper bounds on the attacker errors.
 
For example, suppose we want to audit the Gaussian mechanism.  To compute a lower bound on the privacy parameters of the Gaussian mechanism (i.e, $\mu$), we have:

\begin{align}
    \mu_{emp}^{lower}   = \Phi^{-1}(1-\bar{\alpha})-\Phi^{-1}(\bar{\beta}) \;
\end{align}

We convert this into a lower bound for $\epsilon$ by noticing that the lower bound $\mu_{emp}^{lower}$ implies an upper bound on the trade-off function of the mechanism at every $\alpha$. 
Such an upper bound on the trade-off function enables us to use \Cref{eq:eps_delta_tradeoff} at a fixed $\delta$, to find the largest lower bound for $\epsilon$ over all $\alpha$.

\begin{figure*}
\centering
\begin{subfigure}{0.24\textwidth}
    \centering
    \includegraphics[scale=0.32]{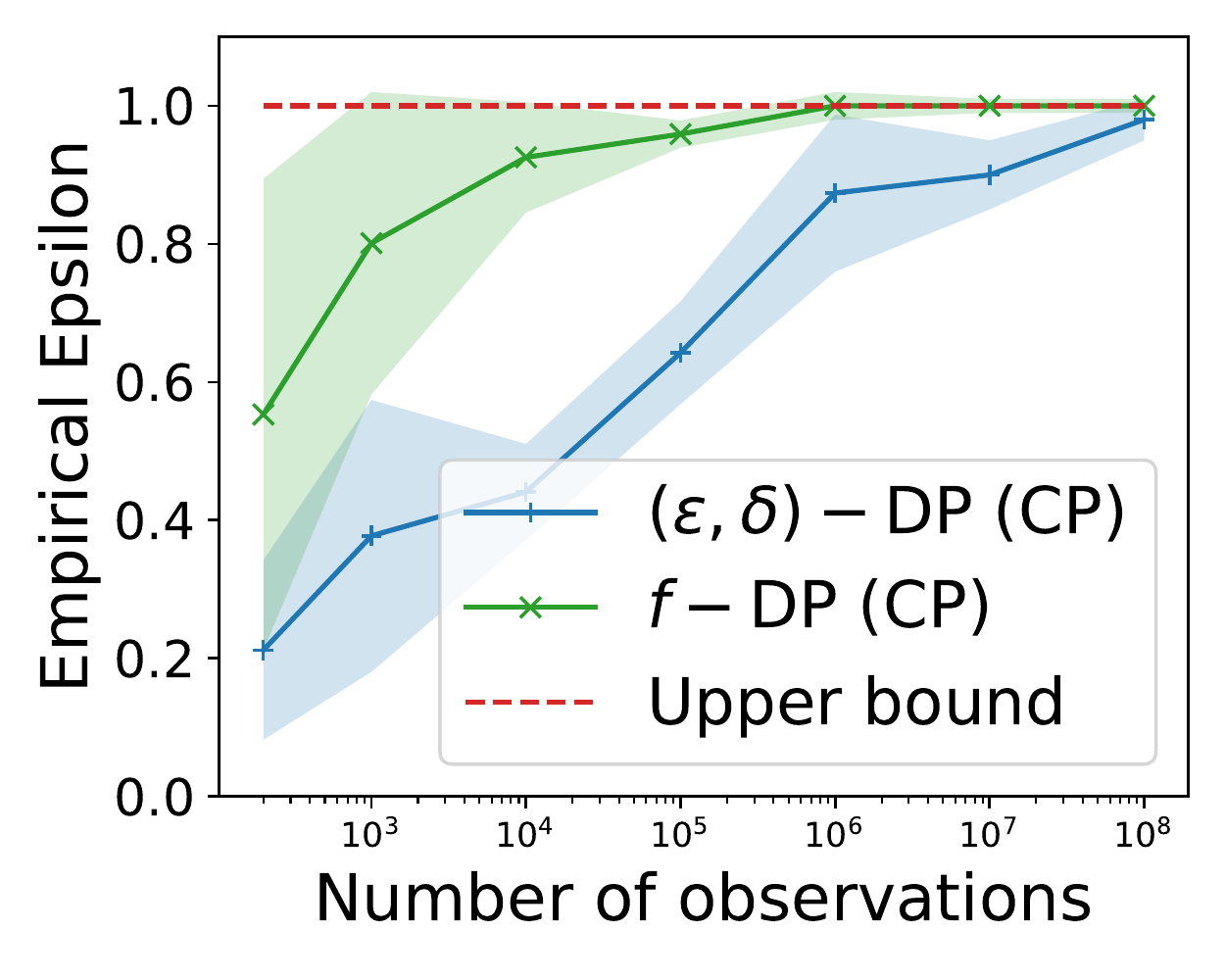}
    \caption{$\varepsilon=1$}
    \end{subfigure}
    \begin{subfigure}{0.24\textwidth}
    \centering
    \includegraphics[scale=0.32]{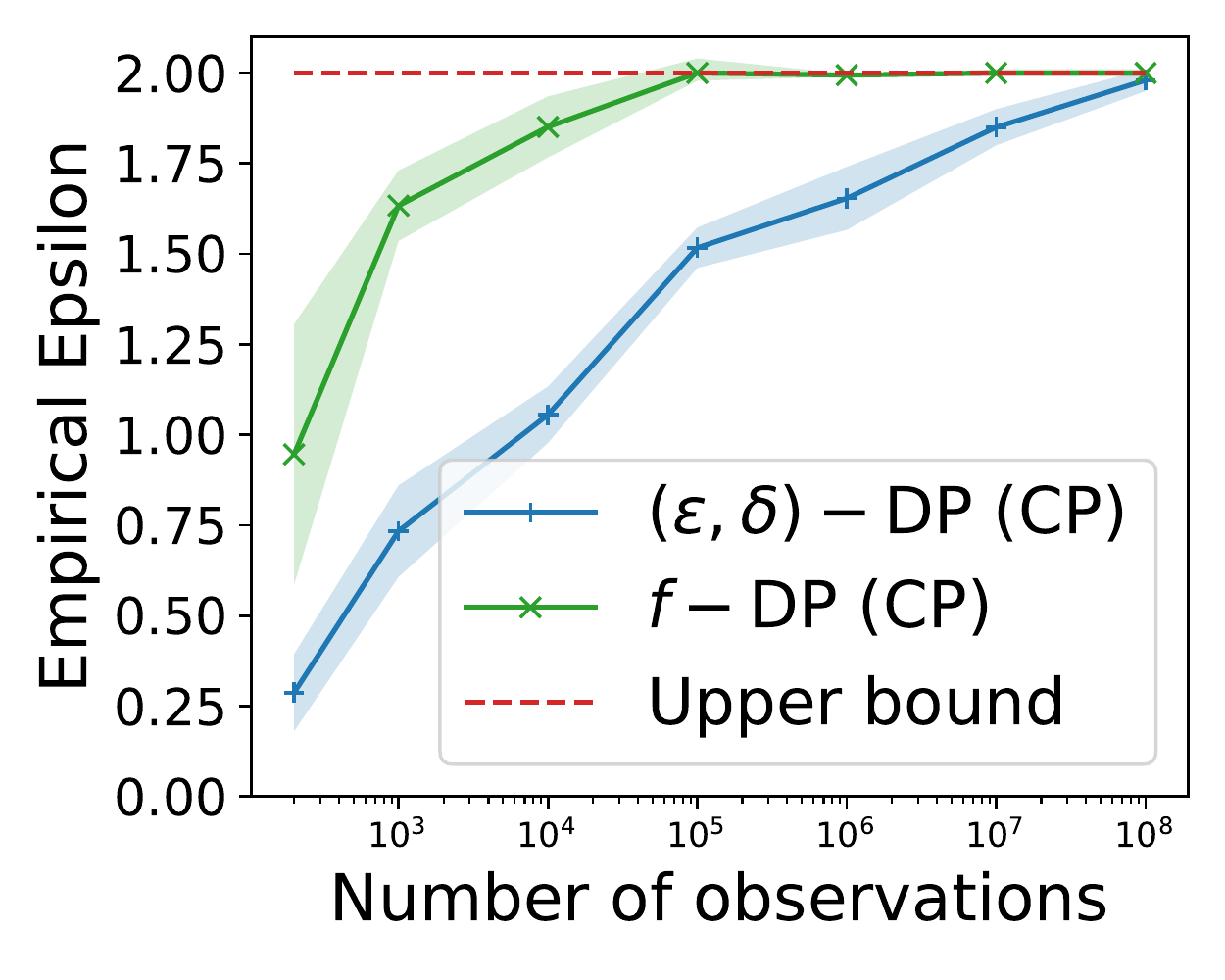}
    \caption{$\varepsilon=2$}
    \end{subfigure}
    \begin{subfigure}{0.25\textwidth}
    \centering
    \includegraphics[scale=0.32]{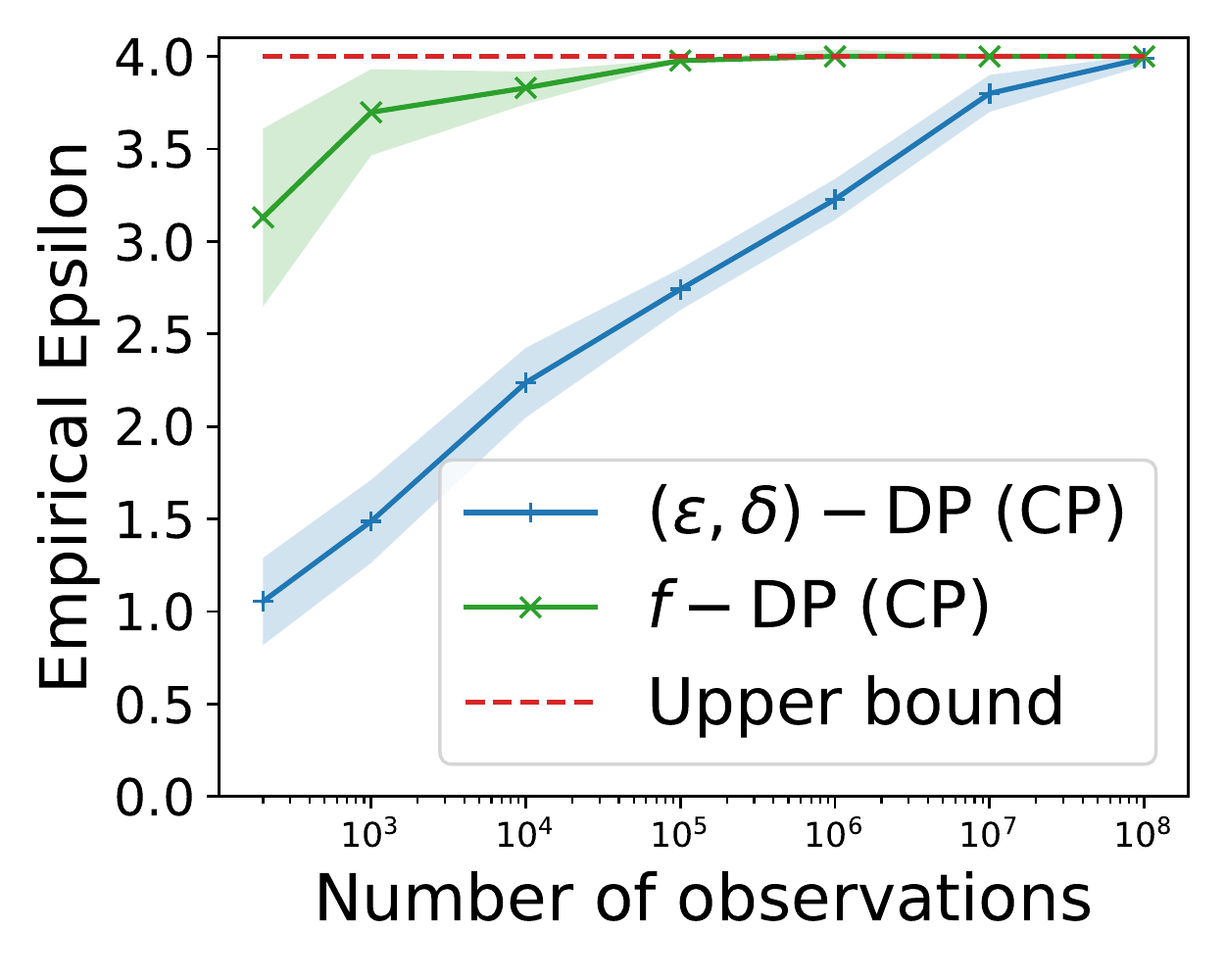}
    \caption{$\varepsilon=4$}
    \end{subfigure}
    \begin{subfigure}{0.24\textwidth}
    \centering
    \includegraphics[scale=0.32]{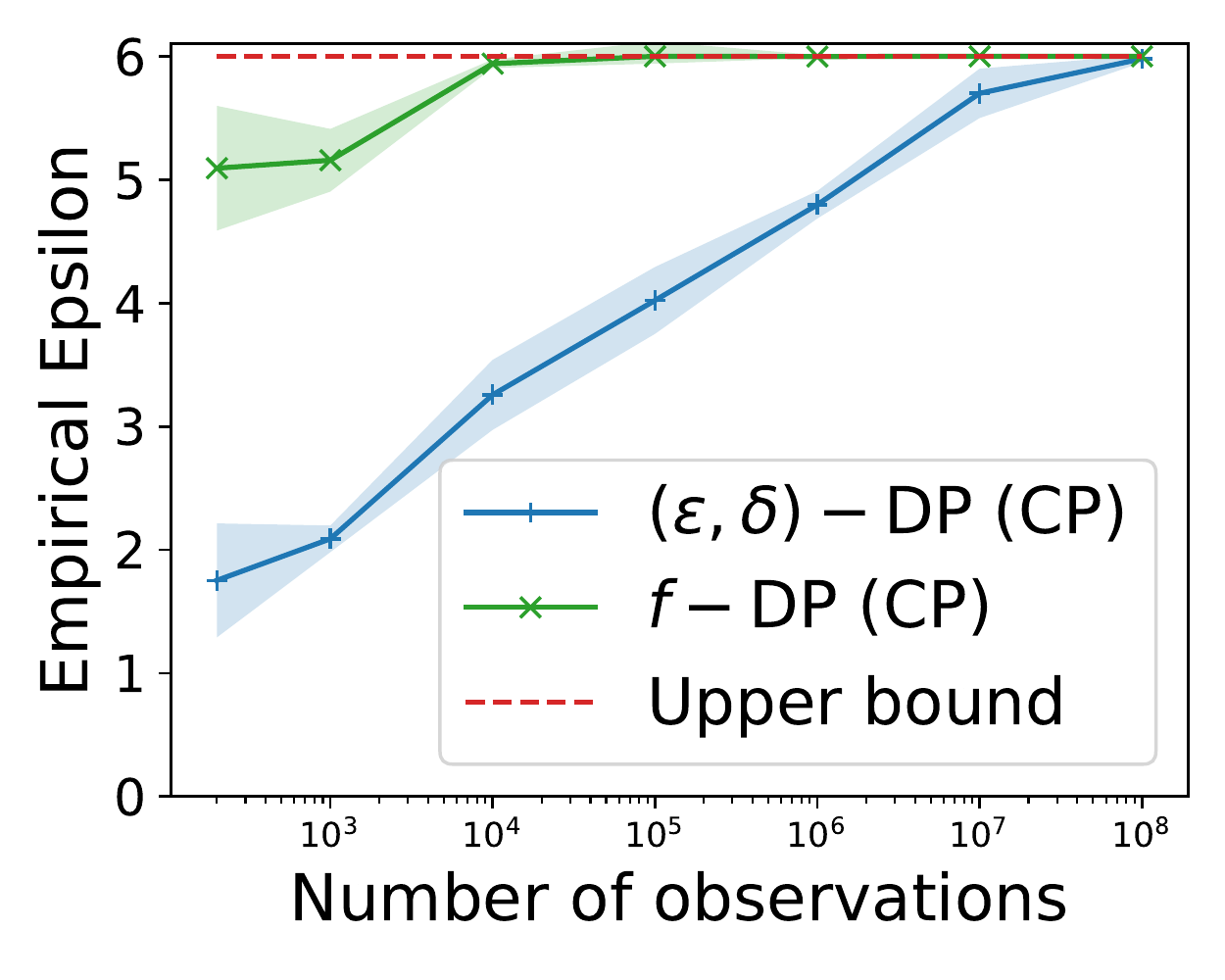}
    \caption{$\varepsilon=6$}
    \end{subfigure}
\caption{Comparison of the \adp definition to audit DP-SGD compared to using \fdp of Gaussian mechanism both converted in \adp and lower bounded using Clopper Pearson (20 independent runs, $\delta=10^{-5}$).}
\label{fig:audit_gdp_dp}
\end{figure*}

\paragraph{Improvement:}
    Nasr et al.~\cite{nasr2021adversary} showed DP-SGD accounting is tight for a worst case dataset ($D=\emptyset$), assuming the adversary has white-box access to all iterations of the training. 
    However, they require many observations to achieve tight bounds, due to the aforementioned drawbacks of auditing with \adp.
    We re-evaluate this setting using our new approach of auditing with GDP. 
    In \Cref{fig:audit_gdp_dp}, we compare the lower bounds found through \adp (\Cref{eq: nasr_lb}) against using the Gaussian trade-off function and converting the Gaussian mechanism parameter to \adp, and we inspect how these two methods compare as the adversary collects more observations from which they compute upper bounds $\bar{\alpha}$ and $\bar{\beta}$.
    While, it is possible for \adp audit to find a lower bound that is tight to the theoretical value for $\epsilon$, this is only achieved when the adversary has 100 million observations. 
    When the number of the observations is smaller there is a non-trivial gap between the theoretical bound and the empirical lower bound.
    Comparatively, if we use the Gaussian mechanism's trade-off function (GDP) to estimate a lower bound on the privacy parameter $\mu$, and convert this into a bound on $\epsilon$, we can achieve a tight estimate even with 1,000 observations. 
    % Moreover, by using the trade-off function of the privacy mechanism instead of the generic \adp the auditing is less sensitive to the decision threshold that the auditor uses to compute the false/true positive rates.
    % Figure~\ref{fig: threshold_comparison_nsamples_5000} shows result of the auditing for different choices of threshold either using GDP or \adp. 
    % As we can see auditing using \adp is much more sensitive to the decision threshold compared to GDP.  
    % We discuss this in more detail in \Cref{app:choose_threshold}.
    %In our experiment section, we show how by using this technique we can achieve tight results in more detail other settings as well.

% \begin{figure}[t]
% %  \captionsetup{width=1.\textwidth, justification=centering}
% \centering
%     \includegraphics[width=1.\linewidth]{figures/compare_gdp_and_cp.pdf}
% \caption{We compare \fdp and \adp with Clopper-Pearson lower bounds for $5K$ observations. The lower bound found through GDP are approximately the same for any decision threshold within the observation's support, whereas the \adp with Clopper-Pearson lower bound varies dramatically. We also visualize observations when the canary was and wasn't included in training as histograms.}
% % we used $\delta=10^{-5}$}\tas{$\delta=10^{-5}$?}
% \label{fig: threshold_comparison_nsamples_5000}
% \end{figure}

\paragraph{Approximating the trade off function:}  While using GDP to audit can give us a tight lower bound on the estimated privacy in worst case settings, analyzing the privacy cost of a complex mechanism such as DP-SGD (that needs both sub-sampling amplification and composition over multiple update steps) is non-trivial. Dong et al.~\cite{dong2019gaussian} suggest to use a Central Limit Theorem for DP to compute a closed form solution, unfortunately, this can lead to significant underestimation of the privacy cost in general settings~\cite{koskela2020computing}. 
Therefore, to analyze DP-SGD we use an empirical approach called the ``Privacy Loss Distribution (PLD)''\cite{koskela2020computing} to approximate the trade-off function, which
computes a tight\milad{exact ?} privacy cost of DP-SGD over multiple update steps. 
We refer to Koskela et al.~\cite{koskela2020computing} for a detailed description of PLD, and will interact with it as a black-box $\varepsilon=f_{\mathcal{M}}(\delta)$ that for private mechanism $\mathcal{M}$ and a given $\delta$ will return the exact theoretical $\epsilon$. PLD does not have a closed-form trade-off function, in Appendix~\ref{app:pld} we explain how we use PLD for auditing.

\subsection{Lower Bounding \fdp With Bayesian Estimation}\label{ssec:baysian}

Recently, Zanella-Béguelin et al.~\cite{zanella2022bayesian} showed it is also possible to compute \emph{credible intervals} for $\epsilon$ using a Bayesian method which can significantly reduce the number of observations to estimate a tight bound. 
Here, we show it is possible to extend their approach to lower bound in \fdp (and then convert to a lower bound for $\epsilon$).

\begin{definition} [Cumulative Distribution Function of $f(\alpha,\hat{.})$-DP] Let $u_{(\text{FPR,FNR})}$ be the density function of the joint distribution of (FPR,FNR). The value of cumulative distribution function of $f(\alpha;\hat{.})$ evaluated at $f(\alpha;.)$ is:
\begin{equation}\label{eq:fdp_privacy}
    P_{\hat{.}}(.) = \int_{f(\alpha;.)}^{1-f(1-\alpha;.)} \int_{0}^1 u_{(\text{FPR,FNR})} (\alpha,\beta) \mathop{\wrt\alpha} \mathop{\wrt\beta}
\end{equation}

\end{definition}

Using \cref{eq:fdp_privacy}, we can find an empirical lower bound for $\mu$ in $\mu$-GDP, and then convert to a lower bound for \adp. 
As shown by Zanella-Béguelin et al.~\cite{zanella2022bayesian}, using the CDF of the private mechanism parameters given the attack observations of we can compute credible intervals over $\epsilon$. 
As a reminder Zanella-Béguelin et al.~defined  $u_{(\text{FPR,FNR})} $ as follows:
\begin{align}
    u_{(\text{FPR,FNR})}(\alpha,\beta) := & u_{(\text{FPR|FP})}(\alpha) u_{(\text{FNR|FN})}(\beta)  &\\
    = &\text{Beta}(\alpha;0.5 + \fn,0.5 +\n - \fn) \times& \nonumber  \\
    &\text{Beta}(\beta;0.5 + \fp,0.5 +\n - \fp) &\nonumber
\end{align}

where $\n\xspace$ is the number of observations used to compute false positive and false negative rates, and $\fn$, $\fp\xspace$ are the total number of false negative and false positives, respectively.

\milad{@borja can you add some explanation why this not valid}
    
% \paragraph{Gaussian Differential Privacy: }
% Given that Gaussian mechanism is one of the most common private mechanism, we instantiate the CDF of Gaussian mechanism as follow:

% \begin{definition}[Cumulative Distribution Function of $\mu'$-GDP]  Let $u_{(\text{FPR,FNR})}$ be the density function of the joint distribution of (FPR,FNR)~\cite{}. The value of cumulative distribution function of $\mu'$-GDP evaluated at $\mu$-GDP is :

% \begin{equation}\label{eq:gdp_privacy}
%     P_{\mu'}(\mu) = \int_{\Phi(\Phi^{-1}(1-\alpha)-\mu)}^{\Phi(\mu-\Phi^{-1}(\alpha))} \int_{0}^1 u_{(\text{FPR,FNR})} (\alpha,\beta) \wrt\alpha\ \wrt\beta
% \end{equation}

% \end{definition}

\section{Auditing Setup}

% \jhnote{I'd like to promote \cref{fig: threshold_comparison_nsamples_5000}, \cref{fig: threshold_comparison_nsamples_and_eps_delta} here or mention the findings in the text as a contribution.}

% The adversary is given samples from two distributions, and is tasked with predicting which samples were generated when the canary point was included in training.
% The aim is to minimise false positives and false negatives when making this decision.
% As such, the adversary must design a decision function from which these metrics are calculated.
% Next, we explain how we create such a decision function that will maximise our computed lower bound.

As we have seen, the choice of framework used for auditing can affect the tightness of our lower bound for $\epsilon$. 
In this section, we describe our auditing procedure for each threat model described in \Cref{sec:threat_model}, and then discuss the effect of different attacker choices---such as attack specific hyperparameters---have on the audit results.

\subsection{Auditing Procedure}\label{ssec:procedure}

As mentioned in \Cref{sec:threat_model}, we consider three main threat models. 
For the black-box setting, we use \Cref{alg:dpsgd_debug_black} which trains $2T$ models on datasets $D$ and $D'=D\cup \{z\}$ where $z=(x',y')$ is the differing example between $D$ and $D'$, which we refer to as the \emph{canary}. 
Then the auditor evaluates the loss on the canary example of each model trained on $D$ and $D'$; using this set of losses, the auditor chooses a decision threshold and computes $\alpha$ and $\beta$.
We refer to the statistics collected by the adversary as \emph{observations}.

\begin{algorithm}
\caption{Black-box auditing for DP-SGD}\label{alg:dpsgd_debug_black}
\begin{algorithmic}
\footnotesize
\State \textbf{Args:}  training dataset $D$, loss function $l$, canary input $(x', y')$, number of observations $T$
	
	\State Observations: $O \gets \{\}$, $O' \gets \{\}$
	\For{$t \in \{T\}$}
	\State $\theta \gets$ DP-SGD on Dataset $D$ 
	\State  $\theta' \gets$ DP-SGD on Dataset $D \cup {(x',y')}$
	\State $\mathcal{O}[t] \gets l(\theta,(x',y'))$
	\State $\mathcal{O}'[t] \gets  l(\theta',(x',y'))$
	\EndFor
	
	\Return $\theta$ , $O$, $O'$
\end{algorithmic}
\end{algorithm}
% \vspace{-0.6cm}
In the white-box setting, the adversary can observe model parameters at each update step. 
% Moreover in some settings the adversary can affect the parameters of the the training. 
We summarize the approach used for auditing in a white-box setting in Algorithm~\ref{alg:dpsgd_debug}, using either canary gradients or canary inputs as described in \Cref{sec:threat_model}. 
At each iteration of DP-SGD, the trainer independently samples two batches of data, $B$ and $B'$.
In the \emph{White-box access with Input Space Canaries} threat model, the trainer creates a canary sample and adds it to $B'$ with probability $q_c$.
%from which the gradient is subsequently computed, and then added to the batch of gradients before being privatized.

In the \emph{White-box access with Gradient Canaries} threat model, the trainer computes the batch of per-example gradients for $B$ and $B'$ and adds a canary gradient into $B'$ with probability $q_c$.
Batch $B$ has been sampled from the original dataset $D$, while $B'$ has been sampled from a modified dataset $D'$.
In both threat models, after the canary is added, DP-SGD proceeds as normal, clipping, aggregating, and noising the gradient sums.
%creates a gradient and insert this into the batch of gradients, after which the entire batch is privatized.
%, and adds a canary sample to $B'$ with probability $q_c$.
%As a result, $B$ is sampled from the original dataset $D$, while $B'$ is sampled from a modified dataset $D'$.
%In the \emph{White-box access with Gradient Canaries} threat model, the trainer creates a gradient and insert this into the batch of gradients, after which the entire batch is privatized.
%In the  \emph{White-box access with Input Space Canaries} threat model, the trainer creates a sample from which the gradient is subsequently computed, and then added to the batch of gradients before being privatized.
For each batch, the trainer computes the dot product between the privatized gradient sum and the canary gradient (or the gradient from the canary input), resulting in a score (which we again refer to as an observation).
The goal of the adversary is to determine whether a batch was drawn from $D$ or $D'$. 
At the end of each run of the algorithm, the trainer produces $2T$ observations, two for each update, and a fully trained model with parameters $\theta$.

In \Cref{alg:dpsgd_debug}, we consider different sampling rates for the canary example and normal training examples to allow the model trainer to evaluate the mechanism at different sampling rates.
If the trainer sets $q_c=1$, the canary is selected in all iterations, and the audit focuses on the privacy mechanism without data sub-sampling. 
This modification allows us to identify bugs in DP-SGD that are not due to batch sampling.

\begin{algorithm}[t]
\caption{White-box auditing for DP-SGD with {\color{ForestGreen} gradient} or {\color{blue} input} space canaries}\label{alg:dpsgd_debug}
\begin{algorithmic}
\footnotesize
\State \textbf{Args:}  training dataset $D$, sampling rate $q$, learning rate $\eta$, noise scale $\sigma$, gradient norm clip $C$, loss function $l$, {\color{ForestGreen} canary gradient $g'$}, {\color{blue} canary input $(x', y')$}, canary sampling rate $q_c$, function \texttt{clip} that clips vectors to max norm $C$, number of observations $T$, number of training iterations $\tau$.
	\vspace{0.1cm}
	\State Observations: $O \gets \{\}$, $O' \gets \{\}$
	\State Trained Models: $\Theta \gets \{\}$
	\State $t \gets 0$
	\While{$t \leq T$}
    	\State Initiate $\theta$ randomly
    	\For {$\tau$ iterations}
    	\State $B_t \gets$ sample instances from dataset $D$ with prob $q$
    	\State $B'_t \gets$ sample instances from dataset $D$ with prob $q$  % 	\tas{TS: should be $D'$ right?}\milad{ we are adding the canary gradient later this should $D$}
    	\State $\nabla [t] \gets \vec{0}$ 
    	\ForAll{$ (x,y)  \in B_t$}
    	\State $\nabla[t] \gets \nabla [t]+ \text{\texttt{clip}}(\nabla_{\theta}(l(x,y))) $
    	\EndFor
    	\State $\widetilde{\nabla[t]} \gets \nabla[t]  + \mathcal{N}(0,\sigma^2\mathbb{I})   $
        \State $\nabla' [t] \gets \vec{0}$ 
    	\ForAll{$ (x,y)  \in B'_t$}
    	\State $\nabla'[t] \gets \nabla' [t]+ \text{\texttt{clip}}(\nabla_{\theta}(l(x,y))) $
    	\EndFor
    	\State $\widetilde{\nabla'[t]} \gets \nabla'[t]  + \mathcal{N}(0,\sigma^2\mathbb{I})   $
    	\State
    	\begingroup
        \color{ForestGreen}
        $g' \gets \text{canary gradient}$
        \endgroup
        or 
    	\begingroup
        \color{blue}
        $g' \gets \text{\texttt{clip}}(\nabla_{\theta}(l(x',y')))$ 
        \endgroup
        \begingroup
    	\State $\widetilde{\nabla'[t]} \gets \widetilde{\nabla'[t]} + g' $ with prob $q_c$ o.w. $\widetilde{\nabla' [t]} + \vec{0}$
    % 	\State $\widetilde{\nabla'[t]} \gets \nabla'[t]  + \mathcal{N}(0,\sigma^2C^2\mathbb{I})   $
    % 	\begingroup
        % \color{ForestGreen}
    	\State $O[t] \gets \langle  g', \widetilde{\nabla[t]}\rangle$
    	\State $O'[t] \gets \langle  g', \widetilde{\nabla'[t]}\rangle$
    % 	\endgroup
    	\State $\theta \gets \theta  - \eta \widetilde{\nabla[t]}$ 
    	\State $t = t+1$
    	\endgroup
    % 	\begingroup
    %     \color{Cerulean}
    %     \State $\theta' \gets \theta  - \eta \widetilde{\nabla'[t]}$ 
    % 	\State $\mathcal{O}[t] \gets l_{\theta}(x',y')$
    % 	\State $\mathcal{O}'[t] \gets l_{\theta'}(x',y')$
    % 	\endgroup
    	\EndFor 
    	\State $\Theta = \Theta + \{\theta\} $
	\EndWhile
	
	\State\Return $\Theta$ , $O$, $O'$
\end{algorithmic}
\end{algorithm}

After collecting observations from a model trained on either dataset $D$ or $D'$, they can be compared with a threshold to compute true and false positive rates.
Next, we will focus on considerations for choosing an appropriate threshold and their effects on the auditing process. We will also discuss other important factors in the auditing process, such as canary selection strategies and the sampling rate $q_c$.

% Here, we discuss the choices in the auditing process that will affect our computed lower bound.

\subsection{Choosing a Decision Threshold}\label{ssec:threshold}

% \jhnote{Change this to refer to observations in the Algorithm instead.}
% The adversary is given samples $X = \{x_1, x_2, \ldots, x_m\}\sim\mathcal{X}$ and $\bar{X} = \{\bar{x}_1, \bar{x}_2, \ldots, \bar{x}_m\}\sim\bar{\mathcal{X}}$, where $\mathcal{X}$ and $\bar{\mathcal{X}}$ represent samples spaces when the canary was and was not used in training, respectively.
% The adversary is tasked with predicting which samples were generated when the canary point was included in training (i.e. correctly attributing $x_i$ and $\bar{x}_i$ to $\mathcal{X}$ and $\bar{\mathcal{X}}$).
% The aim is to minimise false positives and false negatives when making this decision.
% As such, the adversary must design a decision function from which these metrics are calculated.

The output of \cref{alg:dpsgd_debug_black} or \cref{alg:dpsgd_debug} is a set of observations,   with canaries $\{o'_1, o'_2, \ldots, o'_T\}$, and a set without canaries, $\{o_1, o_2, \ldots, o_T\}$.
To compute our attack's FNR and FPR, we must first choose a decision threshold to distinguish between observations from the observation space without the canary, $\mathcal{O}$, or the observation space with canaries $\mathcal{O}'$.

\begin{figure}[t]
%  \captionsetup{width=1.\textwidth, justification=centering}
\centering
    \includegraphics[width=1.\linewidth]{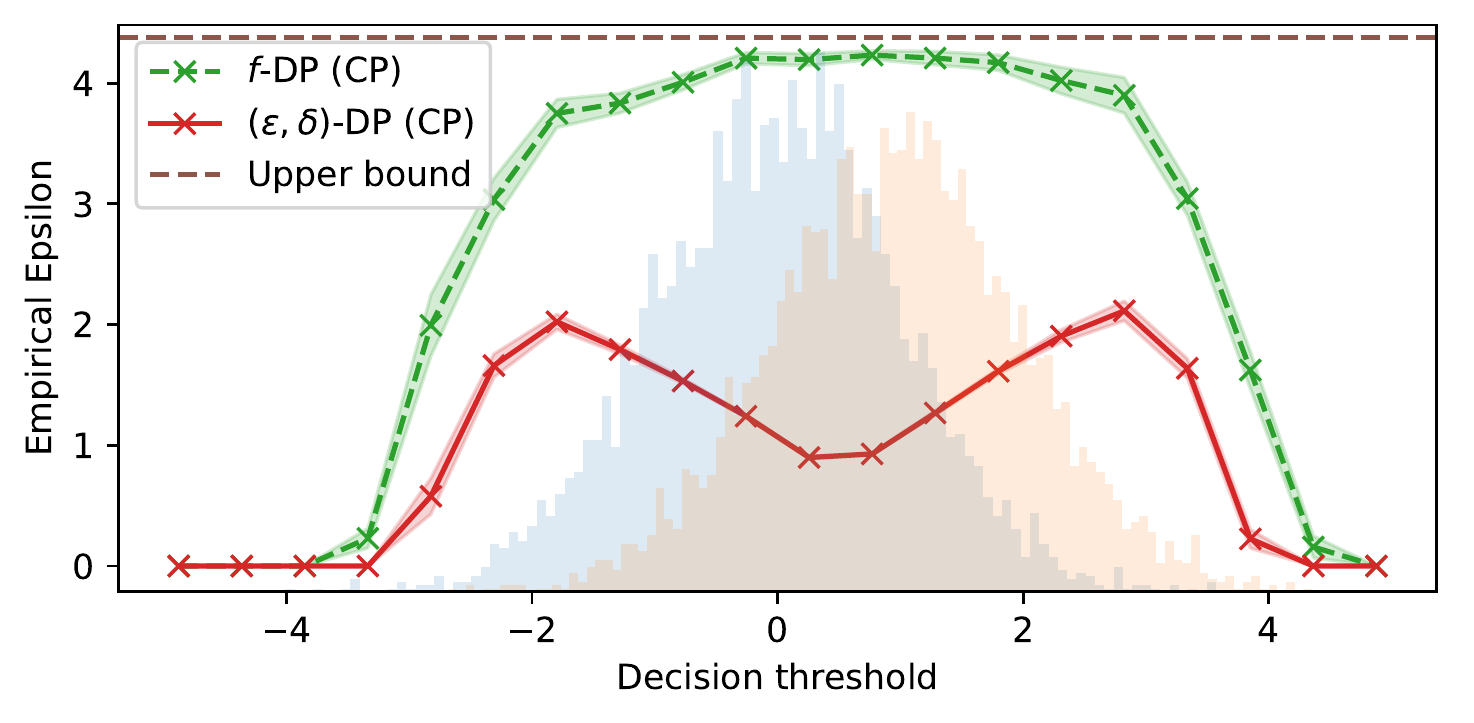}
\caption{An example comparison between \fdp and \adp with Clopper-Pearson lower bounds for 5,000 observations. The lower bound found through GDP are approximately the same for any decision threshold within the observation's support, whereas the \adp with Clopper-Pearson lower bound varies dramatically. We also visualize observations when the canary was and wasn't included in training as histograms.}
% we used $\delta=10^{-5}$}\tas{$\delta=10^{-5}$?}
\label{fig: threshold_comparison_nsamples_5000}
\end{figure}

In the white-box threat model, our observation is $o=\langle  g', \widetilde{\nabla[t]} \rangle$ \at{should the previous be o and not o`?}\jhnote{No i think this is correct, $o$ is the inner product between the canary gradient and the privatized gradient (that didn't include the canary)} and $o'=\langle  g', \widetilde{\nabla'[t]} \rangle$, where $g'$ is the canary gradient, $\widetilde{\nabla[t]}$ is the privatized gradient over a batch $B$, and $\widetilde{\nabla'[t]}$ is the privatized gradient over a batch $B'$.
By construction, we expect $g'$ to be orthogonal to any other gradient $g$ in the batch, $\langle  g', g \rangle = 0$.
In practice, the clipping norm and batch size are known to the adversary and so we can re-scale and normalize the set of observations such that $\mathcal{O}=\mathcal{N}(0, \sigma^2I)$ and $\mathcal{O}'=\mathcal{N}(1, \sigma^2I)$, meaning that in expectation $o$ and $o'$ are sampled from Gaussians with zero and unit mean, respectively.
Thus, the attacker's goal is to distinguish observations sampled from $\mathcal{N}(0, \sigma^2I)$ and $\mathcal{N}(1, \sigma^2I)$ -- we note that this is exactly the same hypothesis testing problem considered in GDP.
% \jhnote{TODO discuss that MSR bounds aren't really bounds}
A benefit of auditing with $q_c=1$ and GDP is that the lower bounds we derive with GDP are agnostic to our choice of decision threshold used to compute type I and type II errors (in the limit of number of observations), while this is not true for \adp. 
This is due to the perfect match between the GDP analysis and the true privacy of the Gaussian mechanism.
% Moreover, by using the trade-off function of the privacy mechanism instead of the generic \adp the auditing is less sensitive to the decision threshold that the auditor uses to compute the false/true positive rates.
\Cref{fig: threshold_comparison_nsamples_5000} shows the result that different choices of threshold have on auditing using either GDP or \adp. 
As we can see, auditing using \adp is much more sensitive to the decision threshold compared to GDP.  
Note that if thresholds we use are found using the same observation data that we compute the lower bound on, the bound is technically not valid.
% However, we compare our default threshold (valid) with these thresholds as it gives an idealised view of what an adversary could feasibly compute.
However, it has become common to report lower bounds on the same set of observations that one uses to find an optimal decision threshold~\cite{zanella2022bayesian, maddock2022canife}, and so for each method we will find the threshold that maximizes the reported lower bound.
% In \cref{tab:threshold_choice}, we survey options for decision thresholds, which we will compare with one another throughout the experimental section.
We stress that for GDP, any decision threshold will be equally likely to maximize the lower bound with a sufficient number of observations; this is not true of \adp.
We discuss this further in  \Cref{app:choose_threshold}.

Next, we discuss how to construct the canary point (either gradient or sample) used in auditing.

\subsection{Canary Type}\label{ssec:canary_types}

The strength of our bound depends on being able to distinguish samples from $\mathcal{O}$ and $\mathcal{O}'$.
In turn, this means crafting canaries that maximize distinguishability.

In the \emph{White-box access with Gradient Canaries} threat model, we use what we refer to as a 
Dirac canary gradient; a gradient with zeros everywhere except at a single index in the gradient vector, where we set its value to the clipping norm $C$.
We compare this choice with other possibilities in \Cref{ssec:canary_grad_choose}.

\begin{algorithm}
\caption{Input canary generation in white-box setting}\label{alg:wb_canary}
\begin{algorithmic}
\footnotesize
\State \textbf{Args:}  In-distribution dataset $D$, model loss function $l$, model parameters $\theta$, $T$ crafting steps, $\eta$ step size
    \State $\vec{g}_{dist} = \frac{1}{|D|} \sum_{(x_i,y_i) \in D} \nabla l(\theta,(x_i,y_i))$
    \State $l_{adv}(x,y) =  |\frac{\nabla l(\theta,(x,y)) . \vec{g}_{dist} }{|\nabla l(\theta,(x,y))||\vec{g}_{dist}|}|$
	\State $(x,y) \sample{D}$
	\For{$t \in \{T\}$}
	\State $x = x - \eta \nabla l_{adv}(x,y)$
	\EndFor
	
	\Return $(x,y)$
\end{algorithmic}
\end{algorithm}
% \vspace*{-0.6cm}
In the \emph{White-box access with Input Space Canaries} threat model, we design a new attack that crafts an input for given model parameters. We evaluate four different canary strategies:   (1) a random sample from the dataset distribution with a wrong label, (2) using a blank sample, (3) an adversarial example, and (4)  and our new canary crafting approach given in~\Cref{alg:wb_canary} and discussed in~\Cref{app:diff_inp}. 
In the \emph{black-box} threat model, we consider a similar range of canary types, which are detailed in \Cref{ssec:investigating_exps}.

\subsection{Canary Sampling Rate ($q_c$)}

\begin{figure}[t]
    \centering
    \includegraphics[width=0.6\columnwidth]{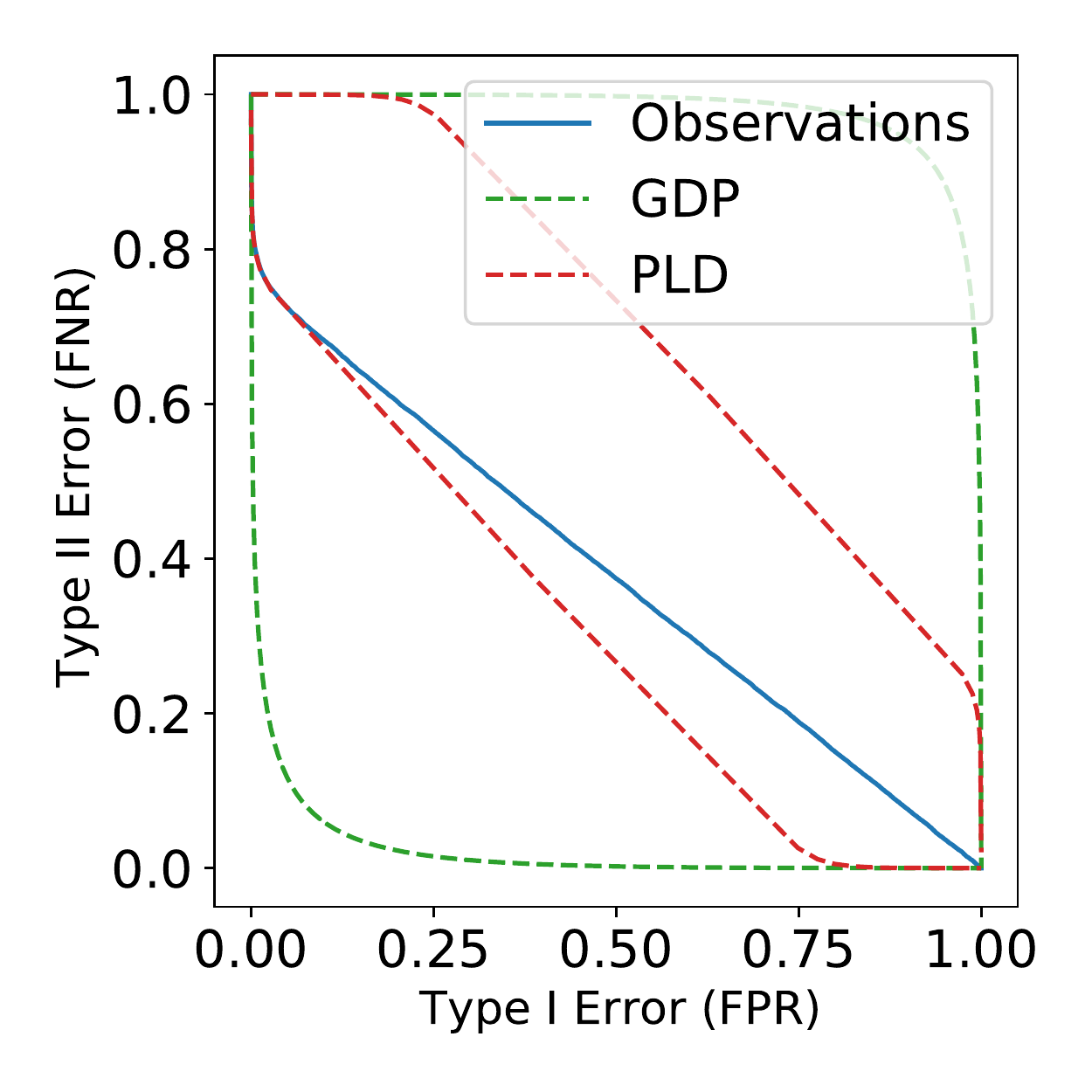}
    \vspace{-0.2cm}
    \caption{PLD or GDP analysis does not describe the achievable error rate from a sub-sampled Gaussian mechanism.}
    \label{fig:sub_sampling_vs}
    \vspace{-0.2cm}
\end{figure}

The analysis of the sub-sampling mechanism in the worst case is tight using \pld/\fdp~\cite{koskela2020computing}. 
From \Cref{fig:pld_apprix_compare}, we see that when $q_c=1$, the observations closely match the theoretical FPR-FNR trade-off.
% curve here we see a large gap between the empirical curve and the theoretical upper bound. This will make auditing the sub-sampled mechanisms hard and they require optimal thresholds and attacks with low false positive rates. 
Instead, in \Cref{fig:sub_sampling_vs} we audit a sub-sampled Gaussian mechanism with sampling rate of $q_c=\frac{1}{4}$, setting $\sigma^2=0.3$ and the number of collected observations to 10,000. 
This figure plots the FPR-FNR curve predicted by using the \pld accounting and also GDP accounting with an equivalent $\varepsilon$ with $\delta=10^{-5}$, and compares it to the empirical curve found through auditing this sub-sampled Gaussian mechanism. 
Clearly, both \pld and GDP upper bounds the observed FPR-FNR curve, but tends to overestimate the trade-off between FPR and FNR, particularly at higher false positive rates~\cite{dong2019gaussian}. 
This suggests that to accurately audit a sub-sampled privacy-preserving mechanism, it may be necessary to use attacks with more precise false positive rates and optimal thresholds in order to achieve tight bounds. 
As mentioned in \Cref{ssec:threshold} (and expanded upon in \Cref{app:choose_threshold}) by using $q_c=1$ (and a sufficient number of observations) we do not need to find the optimal threshold, as the lower bound found using GDP auditing is threshold agnostic, and any threshold will results in tight auditing (which ensures a valid confidence interval and lower bound).

When using auditing to debug an implementation of DP-SGD, we focus on the auditing of the privacy mechanism itself, rather than the sub-sampling process.
We therefore set $q_c=1$ throughout most of the experiments in \Cref{sec:experiments}.
However, we will experiment with the sub-sampled Gaussian mechanism ($q_c<1$) in the black-box threat model, where we are more focused on effect of the threat model on privacy leakage rather than debugging to check if an implementation of DP-SGD is correct. 

\subsection{From Step-wise to End-to-end Auditing}
\milad{@Thomas can you take a look at this }\jhnote{probably move this to somewhere else. Can we argue this is valid because \pld is tight?}\tas{Do we need to use PLD or can we just use GDP accounting? The latter seems easier to justify.}\milad{can we use GDP for the subsampled Gaussian mechanism (this would be more general)? }\tas{If we use subsampling, then we should use PLD/f-DP, but it seems harder to justify than without subsampling. Basically, GDP is making an assumption that the privacy loss is Gaussian, which is a reasonably natural assumption. If we use subsampling, then we are assuming the privacy loss distribution is some weird unnamed distribution of the form $\sum_i \log(1+\exp(X_i))$ where the $X_i$s are independent Gaussians. This is a bit harder to argue is a reasonable assumption.}\milad{in the experiments that we use per step auditing we don't use subsampling. However, if we assume we know the subsampling rate (which is the case in the whitebox), is it still a unreasonable assumption?} \tas{I tried to explain as best I could below. PTAL.}

Our white-box audit observes each step of DP-SGD. As such, we generate privacy lower bounds for individual steps.
We convert these into a privacy lower bound for the end-to-end training procedure by appealing to tight composition results.
For instance, if we assume that each step is tightly characterized by Gaussian DP, then the tightness of Gaussian DP composition allows us to infer that the end-to-end training procedure is also tightly characterized by Gaussian DP and that we can sum up the privacy parameters.
In other words, if the privacy loss distribution of one step is Gaussian, then the privacy loss distribution of the end-to-end procedure is also Gaussian; this is because composition simply adds/convolves the privacy losses. 
Given a lower bound for a single step of DP-SGD, $\epsilon^{\text{lower}}_{\text{emp}}(\delta)$, we find a Gaussian noise scale $\sigma$ that corresponds to this $(\epsilon,\delta)$-DP guarantee. The we compose the corresponding Gaussian DP guarantees to obtain our final estimate.

In DP-SGD, there is also subsampling. That is, we must account for the randomness of the batch selection. In this case, the privacy is not tightly characterized by Gaussian DP.\footnote{Although it is not tightly characterized by Gaussian DP, a subsampled Gaussian can be approximated by Gaussian DP. In practice, this approximation yields conservative estimates of the final $(\epsilon,\delta)$-DP guarantee. Hence this would also be an acceptable auditing methodology.} However, it can be characterized by a more general \fdp guarantee and then we can use \pld to compose over the number of update steps.
In \Cref{sec:experiments}, we show that this method of conversion gives empirical estimates for $\epsilon$ that are close to the end-to-end theoretical $\epsilon$ value.
We note that a similar idea has been explored by Maddock \emph{et al.}~\cite{maddock2022canife}.

\section{Experiments}\label{sec:experiments}

We now evaluate the performance of our proposed auditing technique.
We first demonstrate that auditing with \fdp gives a tight bound on privacy leakage. 
After this, we show that tight auditing can be used for a multitude of purposes, such as investigating if certain choices of training hyperparameters lead to more or less privacy leakage, and debugging implementations of DP-SGD. 

% In the main body of the paper we focus mainly on training models with CIFAR-10. The additional datasets are described in Appendix~\ref{}.
\subsection{Experiment Setup}
We experiment with two commonly used datasets in the privacy literature: CIFAR-10 (with Wide ResNet (WRN-16)~\cite{zagoruyko2016wide} and ConvNet architectures) and Purchase. In addition we also evaluate our experiment on a randomly initialized dataset. In Appendix~\ref{app:exp_details} we describe the hyperparameters and details used to train models. Unless otherwise stated, all lower bounds are given with a 95\% confidence (Clopper-Pearson as in Nasr et al. \cite{nasr2021adversary}) / credible interval (Zanella-B{\'e}guelin et al. \cite{zanella2022bayesian}), and we audit in the \emph{White-box access with Gradient Canaries} threat model using \Cref{alg:dpsgd_debug}. 

% \mcj{does this mean the lower bound has p=0.05 or p=0.025?}\jhnote{Good point, I set p=0.025 for both FPR and FNR and use the union bound to give p=0.05}\milad{same for me}

\noindent\textbf{Terminology:} We describe below the approaches we use to compute lower bounds; our work introduces the $f$-DP strategies.
\textbf{\fdp (CP)}: using the trade-off function of the privacy mechanism with Clopper-Pearson. 
When we do not have the exact trade-off function (e.g, if there are multiple composition of sub-sampled Gaussian mechanisms) we use approximated trade-off function of the privacy mechanism from PLD accounting and Algorithm~\ref{alg:pld_approx}.\textbf{\adp (CP)} \cite{jagielski2020auditing, nasr2021adversary}: using the \adp (\Cref{eq:eps_delta_tradeoff}) trade-off function with Clopper-Pearson. \textbf{\fdp (ZB)} and \textbf{\adp (ZB)} are similar to \textbf{\fdp (CP)} and \textbf{\adp (CP)}, however, we use the Bayesian estimation approach (Section~\ref{ssec:baysian}) to compute lower bounds instead of Clopper-Pearson. \textbf{\adp (ZB)} is equivalent to the approach used by Zanella-B{\'e}guelin et al.\cite{zanella2022bayesian}. \textbf{$\varepsilon-$DP (Katz)} \cite{lu2022general} audits $\varepsilon-$DP with the Katz log confidence interval.

% We divide remaining  experiments into two parts. 

\subsection{(Almost) Tight Auditing of DP-SGD For Natural Datasets Using \fdp}
% All existing works, audit DP-SGD by estimating the end to end privacy cost the learning steps. This approach requires to train many model which is computation very expensive. Because of the heavy computation cost it is only possible to train a limited number of models for small scale models.

Nasr et al.~\cite{nasr2021adversary} showed that DP-SGD is tight with worst-case training sets; however, they observed a noticeable gap between the empirically estimated lower bound and theoretic upper bound for $\epsilon$ when they replace these worst-case datasets with datasets commonly used for DP-SGD benchmarking, even when the adversary has white-box access to the model and can insert canary gradients (e.g., they achieved an empirical $\varepsilon$ lower bound of $<1$ with a theoretical $\varepsilon$ of $8$ on CIFAR-10). 

\begin{figure}[t]
    \centering
    \includegraphics[width=0.85\columnwidth]{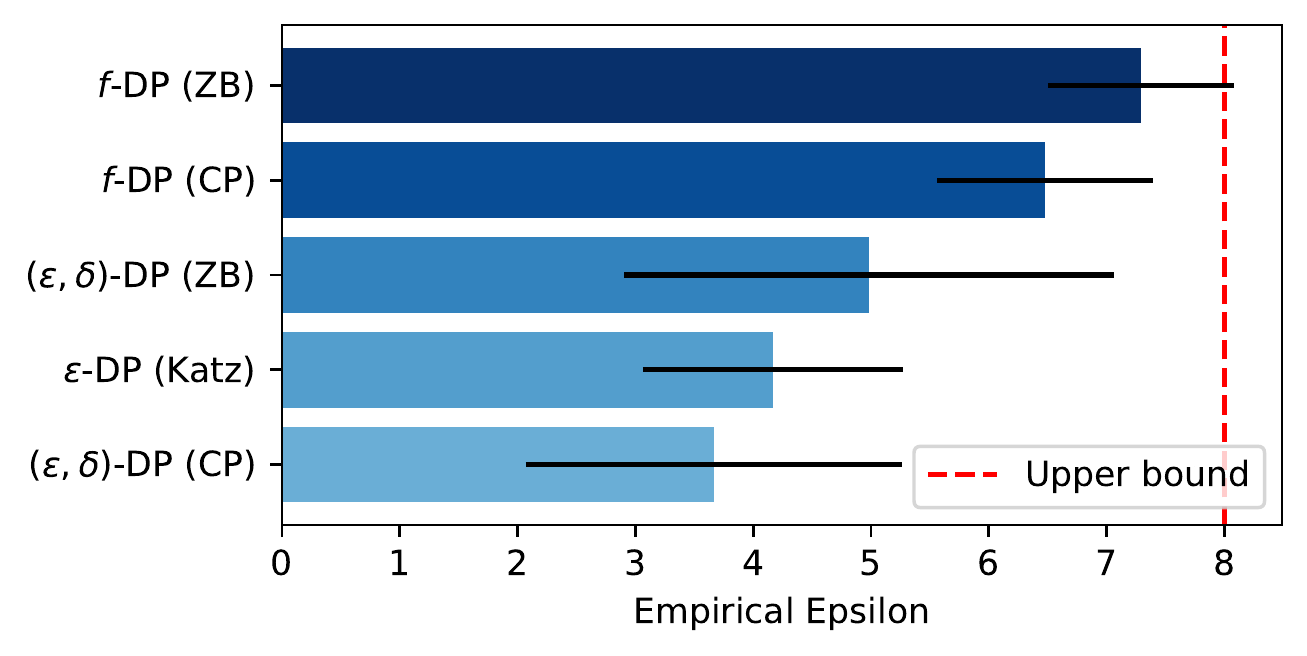}
    \caption{Auditing with \fdp provides the strongest lower bounds with theoretical upper bound $\varepsilon=8$ on CIFAR-10.}
    \label{fig:flagship_cifar10}
    \vspace{-0.5cm}
\end{figure}

We first demonstrate that by auditing with \fdp, we can now compute strong lower bounds with the standard CIFAR-10 training set, where we train and audit a model with 79\% test accuracy at $(\varepsilon=8, \delta=10^{-5})$-DP.
We evaluate and compare our auditing technique against the state-of-the-art auditing methods of Zanella-B{\'e}guelin et al. \cite{zanella2022bayesian} and Lu et al.~\cite{lu2022general}.
Results are given in \Cref{fig:flagship_cifar10}, where we report the average lower bound found over ten independent executions of the experiment along with standard deviation, and the theoretical upper bound for $\epsilon$ given by the privacy accountant.
Regardless of if we use Zanella-B{\'e}guelin et al.'s method for finding credible intervals, or use the Clopper-Pearson confidence interval, the main gain in estimating $\epsilon$ comes from auditing with \fdp.
Auditing with \fdp is almost tight, while the strongest upper bound from prior work is $\sim 5$. 
% Please note we used per step auditing for all of the white-box experiments. 
% \jhnote{Refer to where ever we mention how to convert from per-step to E2E bounds}

\begin{table}[t]
    \centering
    \caption{Comparison of the empirical lower bounds on epsilon with $95\%$ (confidence or credible interval), where the adversary has access to every intermediate model and the adversary can insert a canary gradient vector (white-box setting). We include results on a \emph{Random} dataset---random pixels and labels---of the same cardinality as CIFAR-10.}
    \label{tab:gradient_each_step_all}
    \resizebox{0.49\textwidth}{!}{ 
    \begin{tabular}{@{}lrrrrr@{}}
    \toprule 
    Lower Bounding  & Theoretical $\varepsilon$  & CIFAR-10 WRN-16  & CIFAR-10 ConvNet  & Purchase  & Random  WRN-16  \\
    \midrule
    \multirow{4}{*}{\fdp(CP)}  &  $1$  & $0.75$ & $0.77$ & $0.78$ & $0.74$  \\
    &  $4$  & $3.40$ & $3.34$ & $3.54$ & $3.14$  \\
    &  $8$  & $5.80$ & $6.12$ & $6.40$ & $7.14$  \\
    &  $16$  & $11.14$ & $12.08$ & $12.42$ & $13.14$  \\
    \midrule
    \multirow{4}{*}{\fdp(ZB)}  &  $1$  & $0.95$ & $0.94$ & $0.89$ & $0.90$  \\
    &  $4$  & $3.73$ & $3.80$ & $3.60$ & $3.52$  \\
    &  $8$  & $7.09$ & $7.12$ & $6.94$ & $7.12$  \\
    &  $16$  & $13.95$ & $13.80$ & $13.80$ & $15.14$  \\
    \bottomrule
    \multirow{4}{*}{\adp(CP)}  &  $1$  & $0.41$ & $0.45$ & $0.36$ & $0.35$  \\
    &  $4$  & $1.37$ & $1.80$ & $1.65$ & $1.14$  \\
    &  $8$  & $3.63$ & $3.85$ & $3.25$ & $4.09$  \\
    &  $16$  & $5.25$ & $6.22$ & $6.34$ & $6.96$  \\
    \midrule
    \multirow{4}{*}{\adp(ZB)}  &  $1$  & $0.62$ & $0.62$ & $0.57$ & $0.61$  \\
    &  $4$  & $2.65$ & $2.69$ & $2.45$ & $2.75$  \\
    &  $8$  & $5.07$ & $5.15$ & $4.65$ & $5.09$  \\
    &  $16$  & $5.25$ & $6.22$ & $6.34$ & $6.96$  \\
    \midrule
    \multirow{4}{*}{$\varepsilon-$DP (Katz)}  &  $1$  & $0.49$ & $0.51$ & $0.46$ & $0.41$  \\
    &  $4$  & $1.65$ & $1.95$ & $2.05$ & $2.14$  \\
    &  $8$  & $4.17$ & $3.95$ & $4.24$ & $4.15$  \\
    &  $16$  & $7.52$ & $7.63$ & $7.69$ & $8.01$  \\
    \bottomrule
    \end{tabular}
    }
\end{table}

Our method of auditing with \fdp gives a tight analysis for privacy leakage for both small and large $\epsilon$ and across different datasets (CIFAR-10, Purchase, and a \emph{Random} dataset---random pixels and labels---of the same cardinality as CIFAR-10).
The results, reported in Table~\ref{tab:gradient_each_step_all}, show that our approach \emph{does not require the use of a worst-case dataset to achieve tight estimation of the privacy parameters}.  Lu et al.~\cite{lu2022general} hypothesize that privacy is dataset dependent even in a white-box setting, however, our experiments contradict this hypothesis.  
Given that our results show that tight lower bounds are largely independent of the choice of dataset if the adversary audits in a white-box threat model with canary gradients, our remaining experiments will focus primarily on the CIFAR-10 dataset unless stated otherwise.

We next demonstrate that auditing with \fdp can be useful for detecting implementations of DP-SGD that violate the purported upper bound for $\epsilon$.
As discussed previously, we will concentrate on violations that are not directly caused by sub-sampling, and so our experiments will audit the Gaussian mechanism without composition or sub-sampling, for which we will use the exact trade-off function (GDP instead of the \pld approximation detailed in \Cref{ssec:lb_with_cp}).
% Now that we have established auditing \fdp can find tight bounds is tight for E2E $\varepsilon$ we will report per-step level $\varepsilon$ in following experiments.

% Having shown the auditing with \fdp enables us to tightly characterize the true privacy leakage, we next introduce an example for where this improvement in auditing is useful: debugging implementations of DP-SGD

\subsection{Auditing and Debugging DP-SGD Implementations}

Implementing DP-SGD correctly is notoriously difficult.
Auditing can help identify issues of correctness, as demonstrated by Tram\`{e}r et al.~\cite{DBLP:journals/corr/abs-2202-12219} who used black-box auditing to show the DP-SGD implementation proposed by Stevens et al.~\cite{stevens2022backpropagation} was incorrect and reported a much lower value of $\varepsilon$ than its true privacy leakage.

We investigate how easily our method of auditing with \fdp can detect incorrect implementations of DP-SGD compared to prior work on CIFAR-10.
The upper bound for $\epsilon$ claimed by each DP-SGD implementation throughout the following experiments is 1.27. 
For all experiments in this section we audit a step of DP-SGD, that is, we do not convert our lower bounds into a guarantee of the $\epsilon$ reported after composing across all training steps with \pld.
We do this because even if we were to report a lower bound on the final value of $\epsilon$ (via finding a lower bound for a step of DP-SGD and composing with the (almost) lossless \pld), there will be bugs that cannot be captured by this auditing method.
For example, a bug that is caused by implementing a biased sub-sampling method from the training dataset will likely not be captured by our audit.

\paragraph{Violation 1: Clipping \emph{after} gradient averaging.}

\begin{figure}[t]
%\captionsetup{width=0.9\linewidth, justification=centering}
\centering
\includegraphics[width=0.85\linewidth]{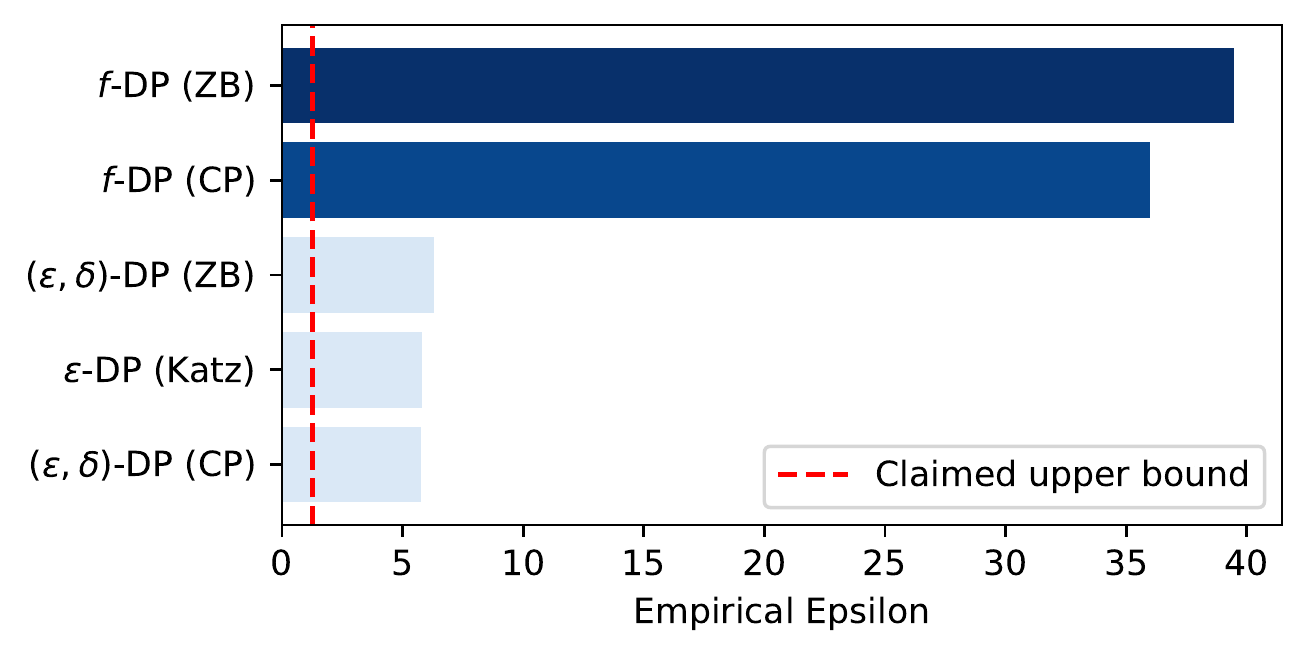}
\vspace{-0.2cm}
\caption{Clipping after gradient averaging bug. We plot the lower bound for $\epsilon$ we can find with each auditing technique when the implementation clips the gradient after averaging in a batch, and so is not \adp with the claimed $\epsilon=1.27$. All methods are able to detect a violation but only \fdp auditing can show the implementation is completely broken, as we can show $\epsilon>35$.\milad{@Jamie can we change to setting where Clopper cannot get a good result }\jhnote{hmm maybe, what settings would expect (eps, delta)-DP + CP to not be able to detect a clipping bug? I kind of like that they all find the bug here but only \fdp shows that the implementation is totally broken, if someone was to use (eps, delta) audit they'd find the bug but incorrectly conclude the mechanism still provides some privacy}}
\label{fig: dp_bugs_no_clip}
\vspace{-0.5cm}
\end{figure}

In DP-SGD, individual gradients must be clipped to a maximum norm $C$ before aggregation.
If the order of these operations is reversed, clipping after aggregation, then the model will not be \adp.
In \Cref{fig: dp_bugs_no_clip}, we see all auditing methods are able to identify a violation as the lower bound found is much larger than the reported upper bound.
However, one may still incorrectly assume that the implementation retains some privacy if we do not audit with \fdp, as the best lower bound we can find is $<10$. 
By auditing with \fdp, it becomes clear that the implementation is completely broken.

\paragraph{Violation 2: Biased noise sampling.}

At every step of DP-SGD, random Gaussian noise must be added to gradients.
If the noise is not randomly sampled then the model will not be \adp.
In practice, we generate a Gaussian noise sample by seeding a random number generator.
We train a model where the seed can only take on 100 different possible values, meaning there are only 100 different possible Gaussian noise vectors.
It may seem that this is a rather contrived example of a DP-SGD bug, but a similar error appeared in the JAX canonical example of how to implement DP-SGD, where a random seed was re-used when adding noise to different sets of model parameters~\cite{githubPrngReuse}.

\begin{figure}[t]
% \captionsetup{width=0.9\linewidth, justification=centering}
  \centering
\centering
    \includegraphics[width=0.85\linewidth]{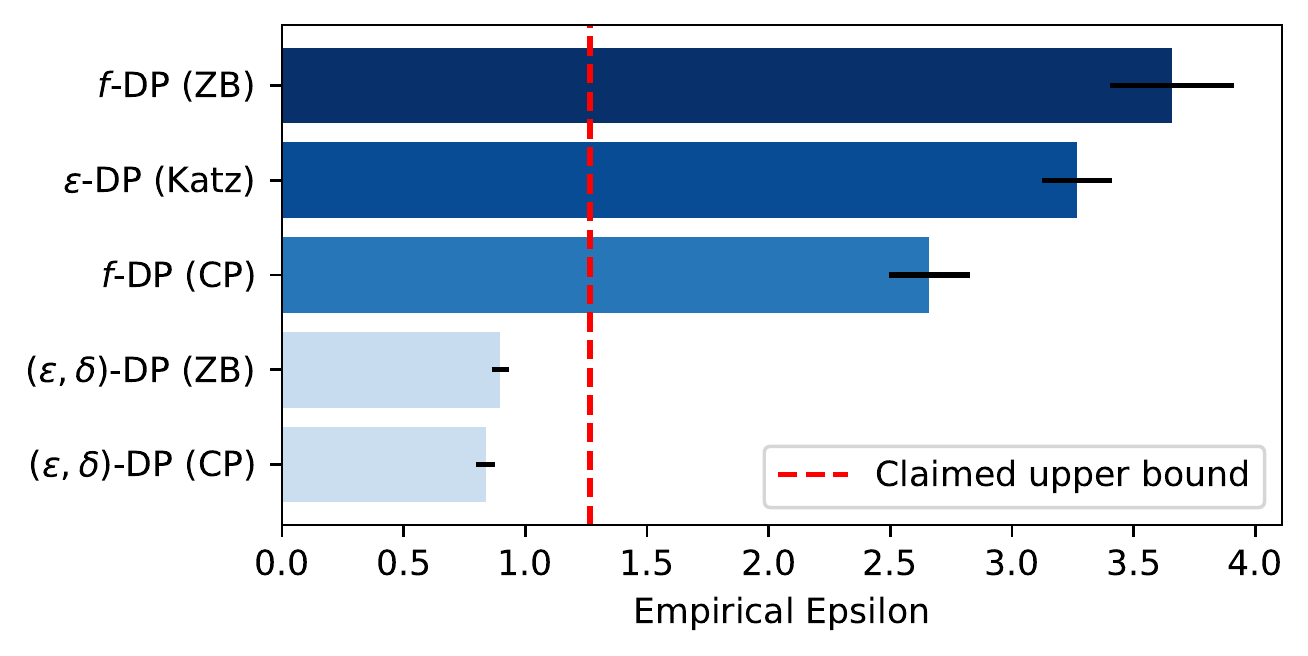}
\caption{Biased noise bug. The Gaussian noise used to privatize gradients in DP-SGD is not sampled randomly. Both \fdp and Lu et al. detect this implementation issue while auditing with \adp fails to detect the issue.}
\label{fig: dp_bugs_biased_noise}
\end{figure}

Results are shown in \Cref{fig: dp_bugs_biased_noise}.
We can detect violations when auditing with \fdp; the auditing method introduced by Lu et al.~\cite{lu2022general} also successfully detects the bug.
However, auditing with \adp directly, either using Clopper-Pearson or the method proposed by Zanella-B{\'e}guelin et al. was not able to identify a violation of the claimed upper bound.

\begin{figure}
% \captionsetup{width=0.9\linewidth, justification=centering}
  \centering
\centering
    \includegraphics[width=0.75\linewidth]{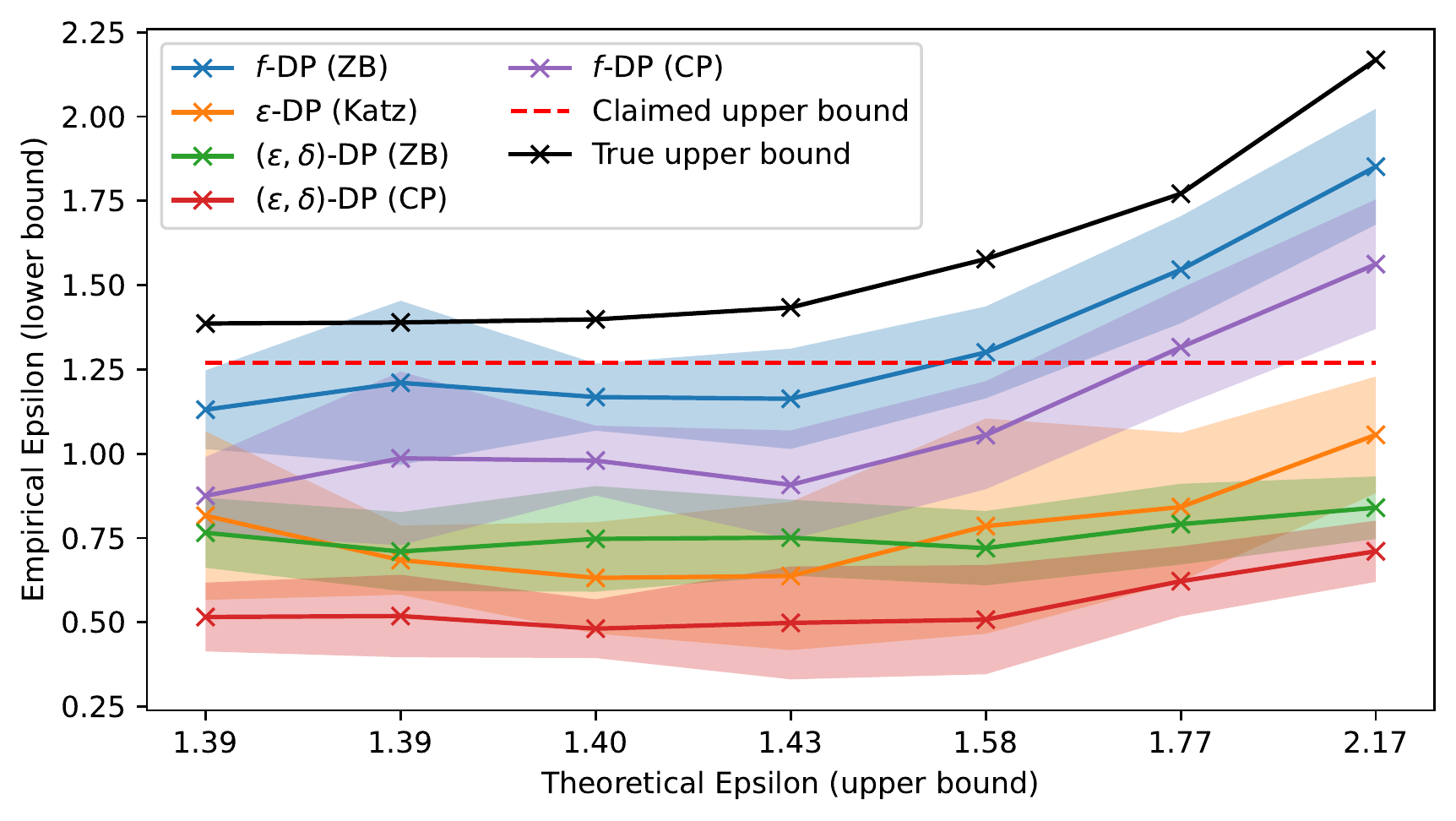}
\caption{Incorrect noise scale bug. We measure how quickly each auditing method can detect a violation of the purported upper bound of $\epsilon=1.27$ when the scale of noise we add to gradients is incorrectly set.}
\label{fig: dp_bugs_noise_too_small}
\vspace{-0.5cm}
\end{figure}

\begin{figure*}[t]
    \centering
    \begin{minipage}{0.65\textwidth}
    \begin{subfigure}{0.49\textwidth}
    \centering
    \includegraphics[scale=0.32]{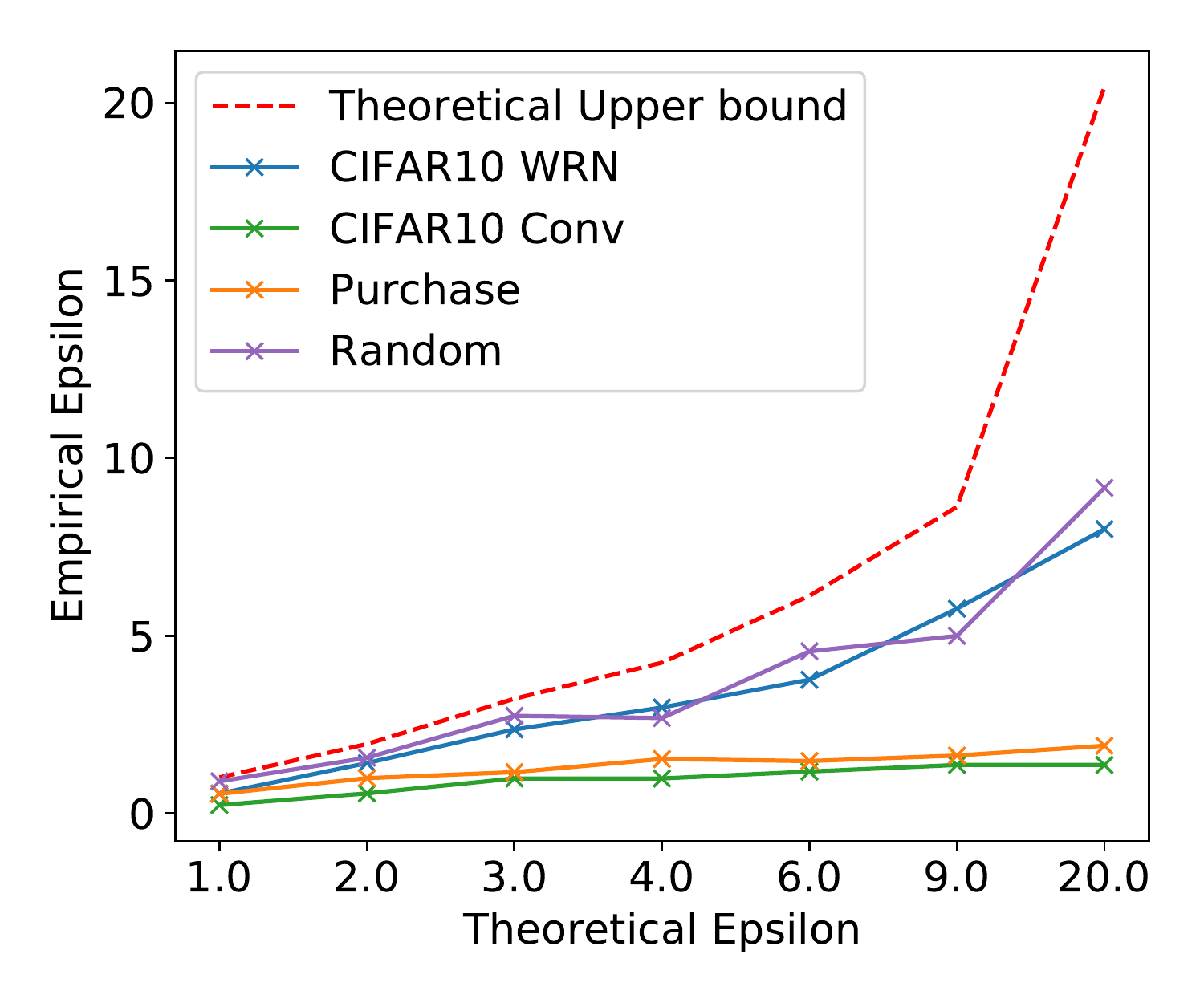}
    \vspace{-0.2cm}
    \caption{\fdp(CP)}
    \end{subfigure}
    \begin{subfigure}{0.49\textwidth}
    \centering
    \includegraphics[scale=0.32]{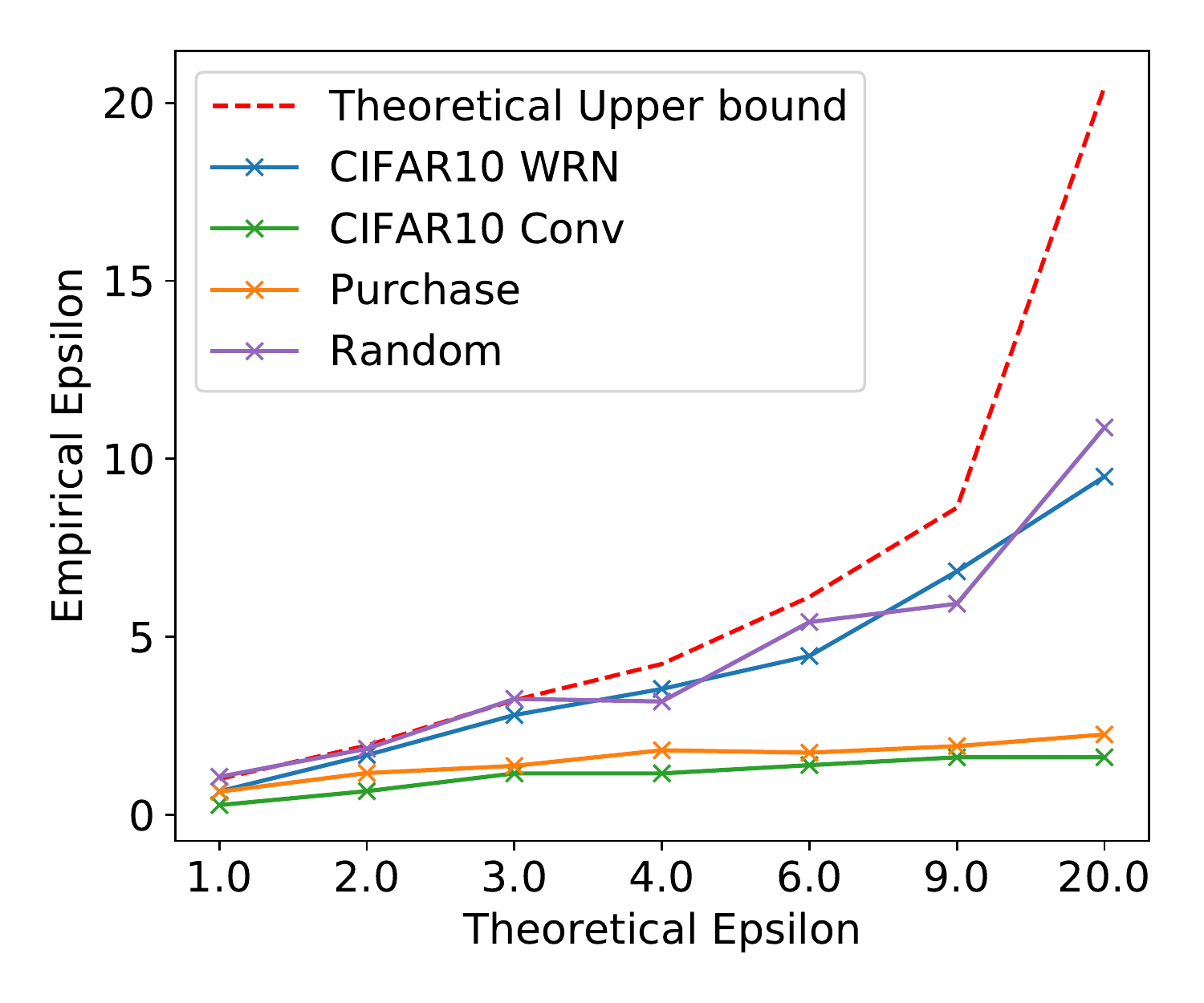}
    \vspace{-0.2cm}
    \caption{\fdp(ZB)}
    \end{subfigure}
    \vspace{-0.2cm}
    \caption{Comparison of the empirical lower bounds on epsilon with $95\%$ (confidence or credible interval), where the adversary has access to every intermediate model and the adversary can insert a canary in input space (white-box setting).}
    \label{fig:input_all_datasets}
    \end{minipage}
    \hfill
    \begin{minipage}{0.33\textwidth}
    \centering
    \captionsetup{type=table}
    \caption{Comparison of the empirical lower bounds on $\epsilon$ (using \fdp (ZB)) in the black-box threat model where the adversary inserts a canary point at the beginning of training, and only gradient clipping is applied on CIFAR-10.}
    \label{tab:e2e_clip_only}
    \resizebox{\columnwidth}{!}{%
    \begin{tabular}{@{}p{4cm}r@{}}
    \toprule 
    Clipping norm $C$ & $\varepsilon$ lower bound \\
    \midrule
    $\hphantom{1}0.1$  &  $23.4$  \\
    $\hphantom{1}1.0$ &  $26.8$ \\
    $10.0$    &  $41.0$   \\ 
    \midrule 
    Maximum $\epsilon$ lower bound at this number of observations & $44.0$ \\
    \bottomrule
    \end{tabular}
    }
    \end{minipage}
\end{figure*}
\paragraph{Violation 3: Noise scale is too small.}

The value of $\epsilon$ is inversely proportional to the scale of noise we add to gradients.
As we decrease the scale of noise, $\epsilon$ increases.
The third bug we investigate is when the noise scale we add is unexpectedly smaller than the target scale we set.
This bug often arises because the the sensitivity (clipping value) needs to calibrated to the batch size in order to compute the correct noise scale, and this is easy to get wrong (c.f. Tram\`{e}r et al. \cite{DBLP:journals/corr/abs-2202-12219}).
For example, in settings where gradient computations are distributed across multiple machines, we could incorrectly add noise to the average gradient found on each machine, and then aggregate. 

We train models with decreasing scales of noise, implying larger values of $\epsilon$ than the claimed upper bound.
Results are shown in \Cref{fig: dp_bugs_noise_too_small}, where we find auditing with \fdp closely follows the true upper bound, meaning we can detect a violation to reported $\epsilon=1.27$ when the true value is $\epsilon=1.57$.
Auditing with \adp directly, either using Clopper-Pearson, Lu et al.~\cite{lu2022general}, or the method proposed by Zanella-B{\'e}guelin et al. does not successfully identify a violation of the claimed upper bound, even when the true upper bound for $\varepsilon$ is as large as 2.17.

\subsection{Investigating Privacy Leakage with Tight Auditing}\label{ssec:investigating_exps}

In this section, we expand our analysis to examine the impact of various settings and parameters on privacy leakage.

\paragraph{Is there a difference between auditing in gradient and input space?}
Auditing with gradient canaries can be useful in specific contexts, such as in federated learning or when debugging a model. However, in most cases, practitioners are more concerned with understanding the effect of a single input space example on the model. This is because this setting more closely measures the privacy leakage that could be experienced by a worst-case training example.

To evaluate how privacy leakage could change by removing the ability to insert a canary gradient, we run experiments in the \emph{White-box access with Input Space Canaries} threat model, as described in \Cref{sec:threat_model}. 
In particular, we use Algorithm~\ref{alg:wb_canary} to create the canaries.
Note that for this experiment we only use the first 250 iterations of DP-SGD to collect observations and estimate the per-step $\epsilon$ lower bound. 
We found that, in this threat model, the first few hundred iterations of DP-SGD leaks more privacy than the entire training run; in other words, the lower bound we compute using the first 250 steps is larger than the lower bound we compute over the entire training run (2,500 update steps).
In general, we find that auditing in the input space becomes weaker if the observations are collected from updates towards the end of training.
We discuss this further in Appendix~\ref{app:diff_steps}.

% In Appendix~\ref{}, we evaluate the effect of different models size and different iterations on the auditing.  

As shown in Figure~\ref{fig:input_all_datasets}, even if the attacker can only insert canary samples (rather than gradients), we can compute tight bounds for $\epsilon<10$. 
We also observe that the choice of the model architecture has an impact on auditing in input space. 
Specifically, we can get an (almost) tight lower bound when using Wide Resnet architecture regardless of the datasets. 
Moreover, when we compare the CIFAR-10 dataset results between Wide Resnet and ConvNet models, we see a large gap which further emphasizes the significance of the model architecture on privacy leakage.
% However, as we show in  Appendix~\ref{}, the fact that we used the first 250 iterations plays an important factor in the tight lower bound and as we use later iterations, the empirical lower bounds becomes looser.

% \paragraph{Do different model architectures lead to different privacy leakage?}

% \paragraph{Does hidden states reduce privacy leakage?}

\begin{figure*}
% \captionsetup{width=1.\columnwidth, justification=centering}
\centering
\begin{subfigure}{0.49\textwidth}
\centering
    \includegraphics[width=.65\linewidth]{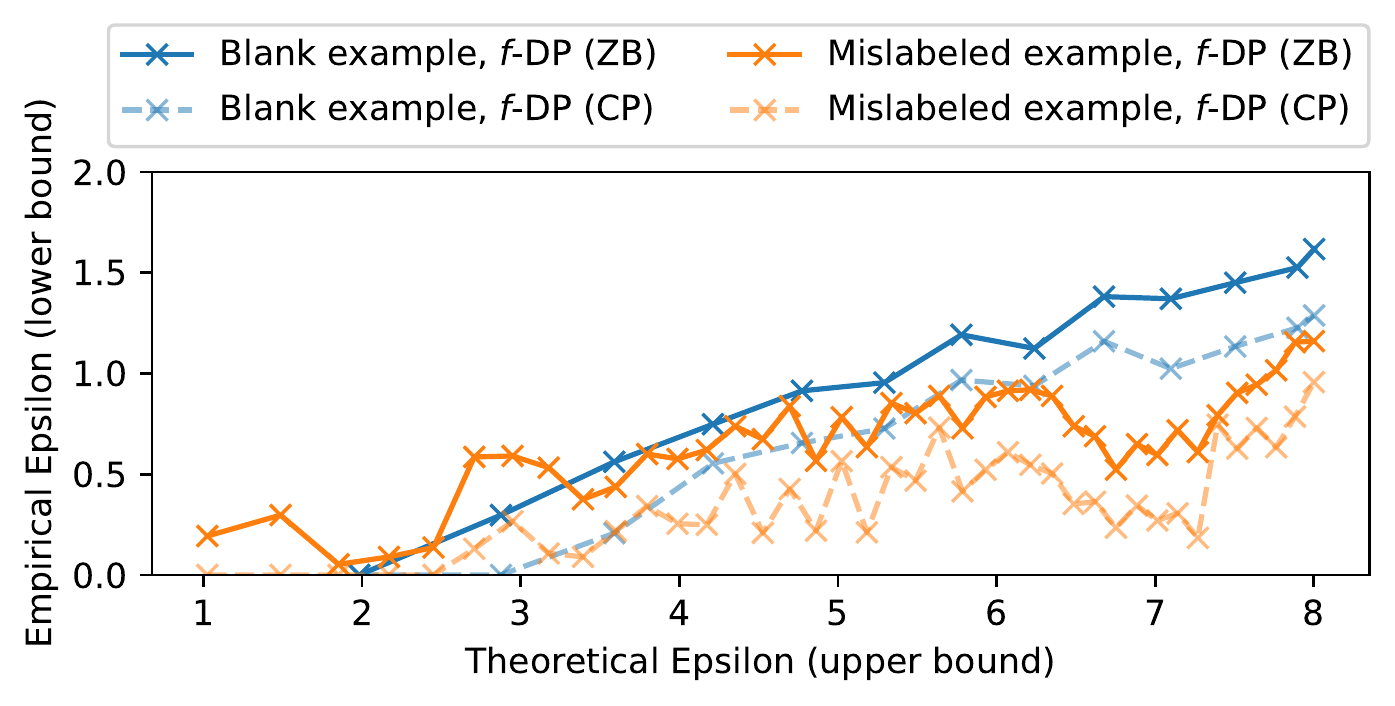}
    \vspace{-0.1cm}
    \caption{CIFAR-10, WRN-16}
\end{subfigure}
\begin{subfigure}{0.49\textwidth}
\centering
    \includegraphics[width=.65\linewidth]{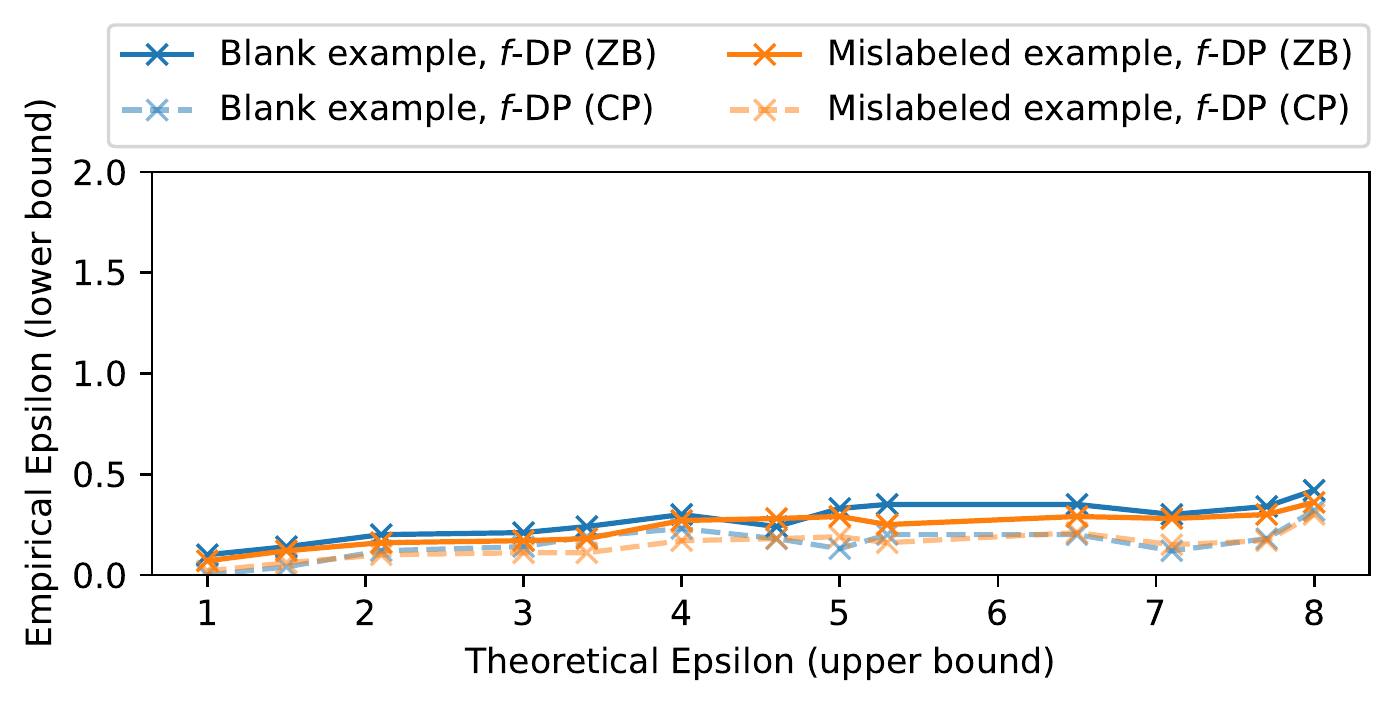}
    \vspace{-0.1cm}
    \caption{CIFAR-10, ConvNet}
\end{subfigure}
\begin{subfigure}{0.49\textwidth}
\centering
    \includegraphics[width=.65\linewidth]{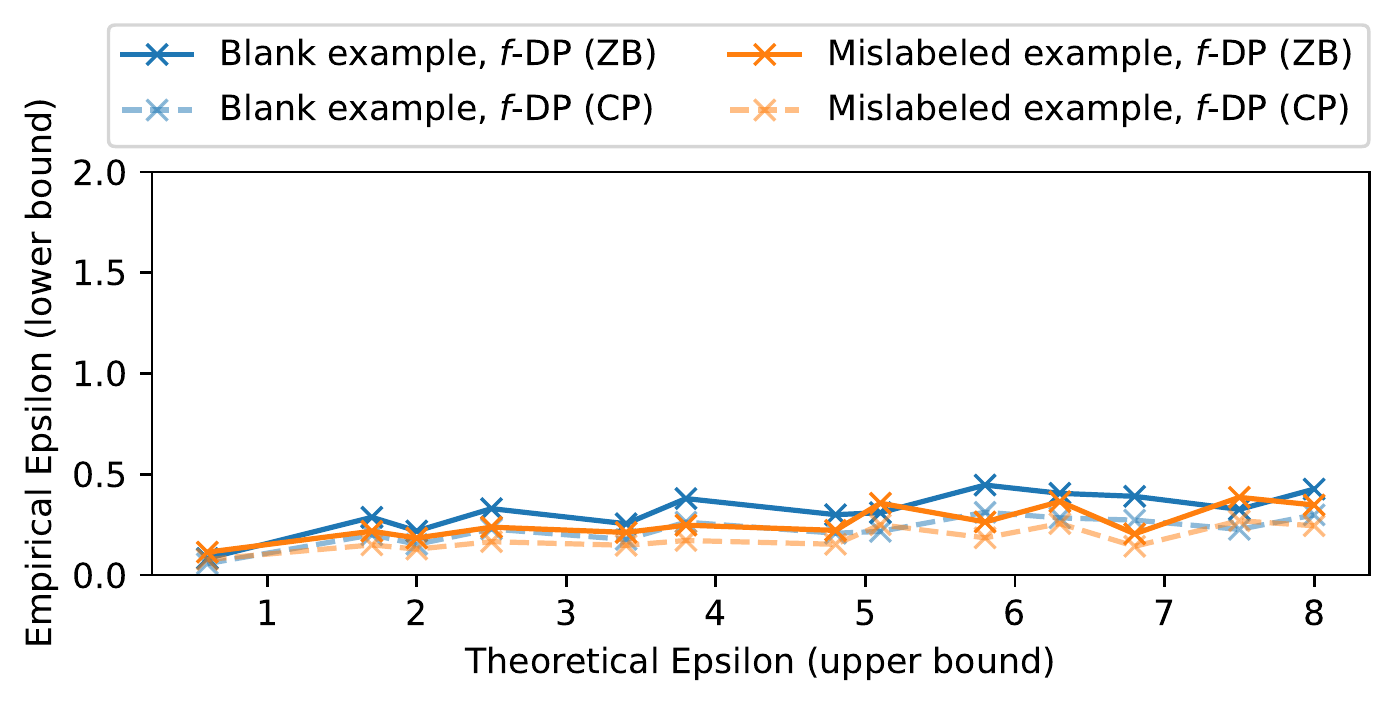}
    \vspace{-0.1cm}
    \caption{Purchase}
\end{subfigure}
\begin{subfigure}{0.49\textwidth}
\centering
    \includegraphics[width=.65\linewidth]{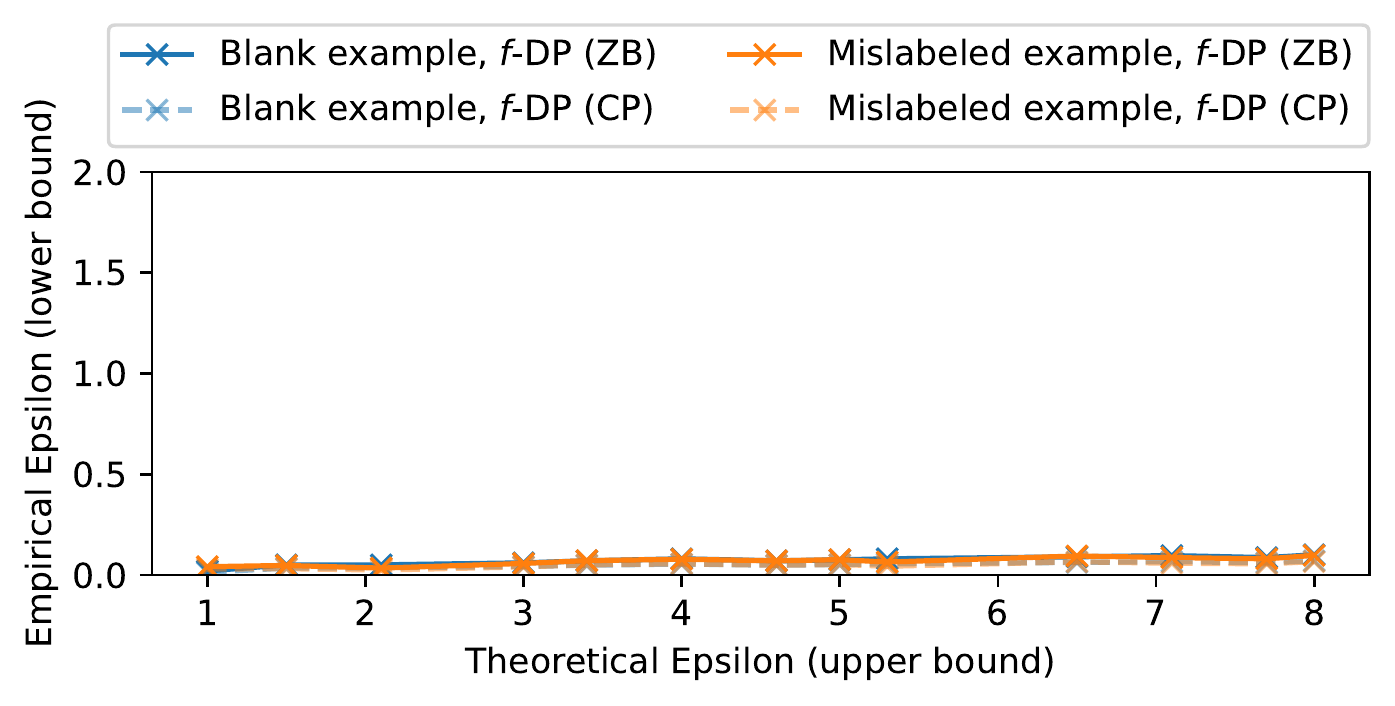}
    \vspace{-0.1cm}
    \caption{Random dataset, WRN-16}
\end{subfigure}
\vspace{-0.2cm}
\caption{Auditing the black-box threat model with \fdp. We train 1,000 models with and without the canary sample that we insert at the beginning of training. We either use a blank or mislabeled image as the canary input.}
\label{fig: e2e_ood}
\vspace{-0.5cm}
\end{figure*}
\paragraph{Does clipping alone help?}

It has been conjectured that clipping individual gradients can provide some privacy even without adding noise \cite{carlini2022membership}.
Technically, these models are not differentially private but we can measure the privacy that clipping provides by computing an $\varepsilon$ lower bound.
Of course, it doesn't make sense to audit clipping alone in a white-box access threat model; no noise is added and so the dot-product value we compute to find type I and type II error rates (as described in \Cref{ssec:procedure} and \Cref{ssec:threshold}) will not be masked by any noise.
Instead, we audit in the black-box threat model by creating a canary point, inserting it into the training set, and then training a CIFAR-10 WRN-16 model.
We then measure the loss after training on the canary image.
We do this 1,000 times when the canary image was included in training and when it wasn't, and record the loss of the canary image in each case. 
We experimented with a range of different canary inputs, but found that blank (white) and mislabeled images produced the strongest lower bounds for $\epsilon$.
In other words, these two canary types had losses that were easily separable depending on if they were included in training or not.

In total, we train 2,000 models: 1,000 when the canary input was in training and 1,000 when the canary wasn't included.
Our results are shown in \cref{tab:e2e_clip_only}, where we see that clipping alone provides no privacy. 
The lower bounds for a clipping norm of 10 are close to the maximum possible lower bound at this number of observations.

\paragraph{How much privacy is leaked in a black-box threat model?}

To measure the impact of switching to a black-box threat model on privacy leakage estimation, we follow a similar auditing procedure as set out by Nasr et al.~\cite{nasr2021adversary}.
We select a canary example,  then
we train 1,000 models with the canary point (+ training set), and 1,000 models with only the training set (canary excluded). 
We then measure the $\log(\frac{p}{1-p})$ for each model, where $p$ is the probability of the canary point with respect to its label.
We take the distribution of $\log(\frac{p}{1-p})$ when the canary point was and wasn't in the training set, and compute $\epsilon$ lower bounds using our \fdp method.

Results are shown in \cref{fig: e2e_ood}, where all models are trained up to $\epsilon=8$.
On CIFAR-10 with a Wide ResNet architecture we are able to find a lower bound of $\sim 1.6$ using \fdp (ZB) auditing.
However, it is difficult to separate the effects of the black-box threat model from the effect that sub-sampling has on privacy leakage estimation, as we saw in \Cref{fig:sub_sampling_vs}, both GDP and \pld tend to overestimate the observed trade-off between type I and type II errors in the sub-sampled Gaussian mechanism (with composition over multiple iterations). Additionally, the auditing results on other datasets/architectures are significantly lower compared to those obtained with the Wide ResNet architecture on the CIFAR-10 dataset. Specifically, when examining the random dataset, no non-trivial lower bounds for auditing can be achieved. One of the key factors that distinguishes these experiments is the final accuracy the model is able to attain. 
At $\epsilon=8$, the Wide ResNet architecture for CIFAR-10 is able to achieve a test dataset accuracy of greater than 70\%, compared to less than 50\% for the ConvNet architecture and less than 50\% for the Purchase dataset. Our results suggest that there remains a gap between the theoretical privacy upper bound and the empirical lower bound that can be achieved in a black-box setting, but the size of this gap is highly dependent on the dataset and model architecture.

% \jhnote{Run with other OOD types?}

\section{Conclusion}
As differentially private machine learning becomes more popular and fewer expert users begin to implement such methods, the possibility of bugs and implementation errors will increase.
We provide a simple auditing technique that can achieve tight estimates of privacy leakage on standard ML benchmark datasets.
Our method can be easily integrated with privacy preserving libraries (TF-privacy, Opacus, JAX privacy) to give online estimation of the private mechanism parameters and provide an empirical test for the assumed privacy budget, and only increases the computational overhead by a factor of two.

\bibliographystyle{acm}
\bibliography{refs}

\appendix

% \section{Algorithms}
% In this section, we provide the details on the algorithms used in this work.
\section{Using PLD to approximate the trade off function}\label{app:pld}
\begin{figure}[t]
    \centering
    \includegraphics[scale=0.4]{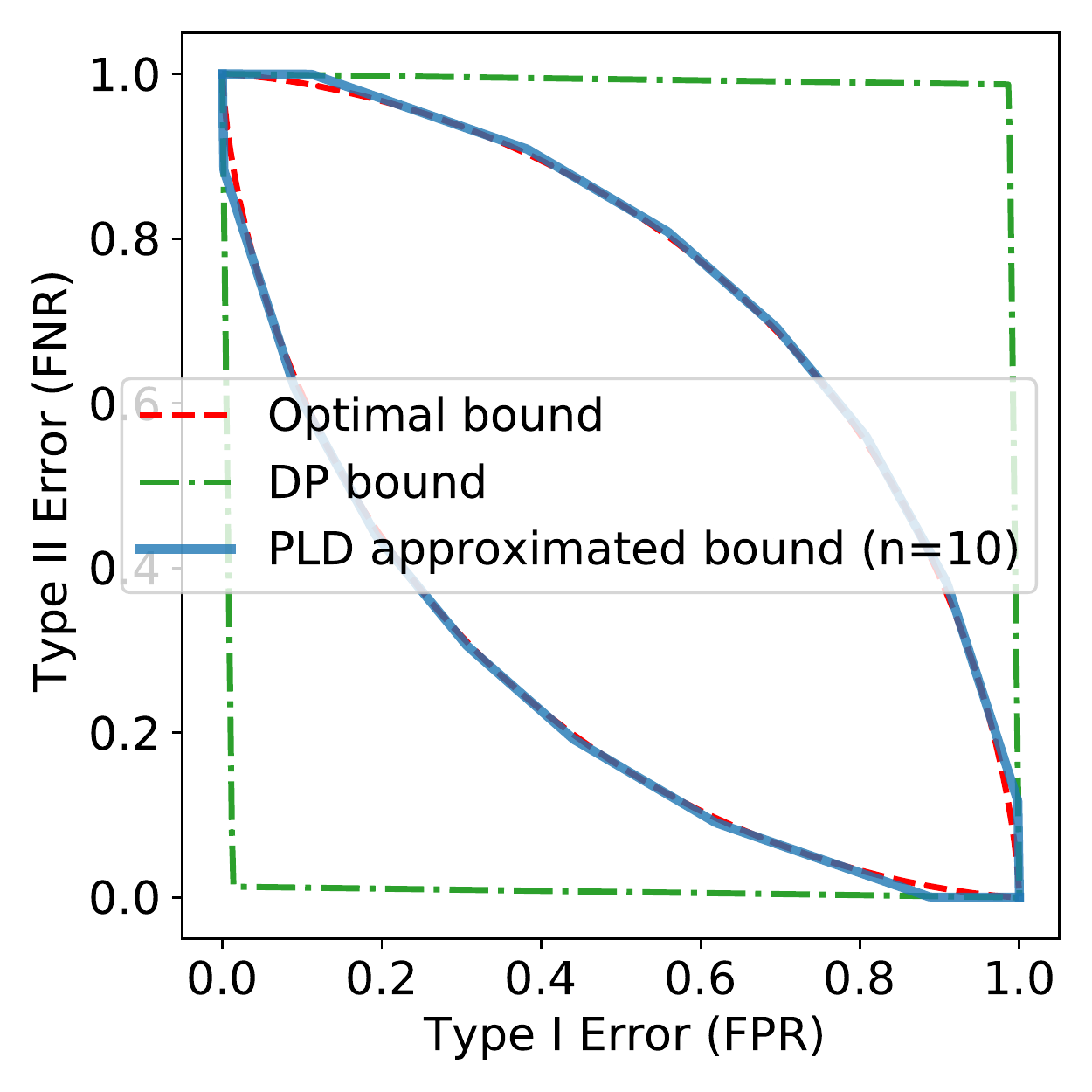}
    \caption{Comparison of the trade-off functions when using an approximation of PLD compared to the optimal curve.}
    \label{fig:pld_apprix_compare}
\end{figure}

PLD does not have a closed-form trade-off function, but we can evaluate empirically a lower bound for a given FPR. 
We approximate the trade-off function using Algorithm~\ref{alg:pld_approx}, which will give us a looser bound for $\epsilon$, however, the difference is in the order of $10^{-1}$.  
Figure~\ref{fig:pld_apprix_compare} compares the approximation approach in Algorithm~\ref{alg:pld_approx} to the optimal trade-off function, as we can see, even with 10 approximations we can get a good estimation of the trade-off function. 
In \Cref{sec:experiments}, we will present experiments that provide lower bounds for a single step of DP-SGD and multiple steps, they will use the GDP and PLD formulations, respectively.
%\todo{Jamie: Some of the more important algorithsm are in the appendix. Maybe we should have an algorithm section in the appendix and put them all there?}

\begin{algorithm}[H]
\caption{Approximating a lower bound on trade-off function using PLD}\label{alg:pld_approx}
\begin{algorithmic}
\State \textbf{Args:} $f_{\mathcal{M}}$ privacy analysis function (outputs $\epsilon$ for a given $\delta$), $n$ number of approximation lines, $\delta$ target delta in privacy analysis
\State $\Delta \gets n$ linearly spaced points between $[ \delta , 1-\delta ] $
\For {$\delta' \in \Delta $ }
\State $\hat{\varepsilon} \gets f_{\mathcal{M}}(\delta')$
\State $l_{\delta'}(x):= \max(0, 1-\delta'-(x  e^{\hat{\varepsilon}}), e^{-\hat{\varepsilon}(1-\delta'-x) } ) $
\EndFor 
\State $l(x) := \min_{\delta' \in \Delta} l_{\delta'} (x)$ \\
\Return $l$
\end{algorithmic}
\end{algorithm}

\section{Experimental Setup}\label{app:exp_details}

On CIFAR-10, we train a Wide ResNet (WRN-16) model~\cite{zagoruyko2016wide} for 2,500 steps with a batch size of 4,096. This reaches approximately 79\% test accuracy at $(\varepsilon=8, \delta=10^{-5})$-DP, which is on a par with state-of-the-art~\cite{de2022unlocking}. We also use a ConvNet architecture with batch size set to 256 and 20,000 steps, which achieves approximately 60$\%$ test accuracy at $(\varepsilon=8, \delta=10^{-5})$-DP. To evaluate the effect of the choice of dataset on privacy leakage, we also experiment with a randomly initialized dataset with the same dimensionality and cardinality as the CIFAR-10 dataset but with pixel values and labels randomly assigned with Wide ResNet architecture. 
For the Purchase dataset we use a three layer fully connected network.

For each dataset in the white-box threat model, we train two models, one without canaries and one where the canary sampling rate is $q_c$.
Unless otherwise stated, we set $q_c=1$.
Of course, we could increase the number of training steps to create a larger number of observations, but we found our method is strong enough to produce tight bounds for large $\varepsilon$ with as little as a few hundred observations (as shown in Figure~\ref{fig:audit_gdp_dp}).

\section{Effect Of Auditing Choices}

\subsection{How to choose a decision threshold given a set of observations}\label{app:choose_threshold}

Our gradient-based auditing method will generate data from two Gaussian distributions: $\mathcal{O}=\mathcal{N}(0, \sigma^2I)$ and $\mathcal{O}' = \mathcal{N}(1, \sigma^2I)$.
% \mcj{won't the standard deviations be $\sigma$/clipping norm?}\jhnote{Good point, I'm going to remove this bit because what we rescale by is a practical thing, and theres a mismatch between theory and practice (shuffle vs random sampling data), and its probably going to confuse}
We must choose a threshold from which we predict if observations are from $\mathcal{O}$ or $\mathcal{O}'$. 
This threshold should be chosen on another set of held-out set of observations in order for our lower bound to be valid. 
That is, we cannot choose the threshold that is dependent on the observations from which we compute the lower bound.
We discuss how this threshold should be chosen if one does not have access to a hold-out dataset of observations (e.g. because generating them is prohibitively expensive) but we want our bounds to remain valid.

As a reminder, Nasr et al.~\cite{nasr2021adversary} compute the following lower bound:

\begin{align}
    \epsilon \geq \max\left[\ln(\frac{1-\delta-\alpha}{\beta}), \ln(\frac{1-\delta-\beta}{\alpha}),0\right]
\end{align}
 
 where $\alpha$ and $\beta$ are the false positive and false negative rate of the audit.
 The false negative rate ($\beta$) is given by $\Phi(\frac{z-c}{\sigma})$ and the false positive rate ($\alpha$) is given by $1-\Phi(\frac{z}{\sigma})$. 
 Because the $\epsilon$ lower bound is  symmetric around the decision threshold $\frac{c}{2}$, we can consider only positive decision thresholds $z>\frac{c}{2}$, where $c=1$. 
 For $z>\frac{c}{2}$ the false positive rate $\alpha$ decreases at a faster rate than the false negative rate $\beta$, and so
 $\max\left[\ln(\frac{1-\delta-\alpha}{\beta}), \ln(\frac{1-\delta-\beta}{\alpha}), 0\right] = \max\left[\ln(\frac{1-\delta-\beta}{\alpha}), 0\right]$.

Plugging in $\beta = \Phi(\frac{z-c}{\sigma})$ and $\alpha = 1-\Phi(\frac{z}{\sigma})$ gives:

\begin{align}
\epsilon \geq \max\left[\ln(\frac{1-\delta-\Phi(\frac{z-c}{\sigma})}{1-\Phi(\frac{z}{\sigma})}), 0\right] 
\end{align}

Since $\ln(\cdot)$ is monotonically increasing we can find the maximum lower bound for $\epsilon$ by upper bounding:

\begin{align}
h(z) = \frac{1-\delta-\Phi(\frac{z-c}{\sigma})}{1-\Phi(\frac{z}{\sigma})} 
\end{align}

Setting $h'(z) = 0$ gives:

\begin{align}
\delta = \Phi(\frac{c-z}{\sigma}) - \frac{\phi(\frac{c-z}{\sigma})}{\phi(\frac{-z}{\sigma})}\Phi(\frac{-z}{\sigma}) \qquad\qquad \label{eq:starstar}
\end{align}

where $\Phi'(\cdot) = \phi(\cdot)$. Let $w$ denote the value of $z$ that satisfies \cref{eq:starstar}. Then the maximum $\epsilon$ lower bound is given by:

\begin{align}
& \ln\bigg(\frac{1-(\Phi(\frac{c-w}{\sigma}) - \frac{\phi(\frac{c-w}{\sigma})}{\phi(\frac{-w}{\sigma})}\Phi(\frac{-w}{\sigma}))-\Phi(\frac{w-c}{\sigma})}{1-\Phi(\frac{w}{\sigma})}\bigg) \\
&= \ln\bigg(\frac{\phi(\frac{c-w}{\sigma})}{\phi(\frac{-w}{\sigma})}\bigg) \\
&= \frac{1}{2\sigma^2}(w^2 - (w-c)^2) \\
&= \frac{c}{2\sigma^2}(2w - c)
\end{align}

This is positive because we know $w>\frac{c}{2}$.
Importantly, the maximum lower bound is only achieved if we choose a decision threshold $w$ satisfying \cref{eq:starstar}, which only has a single solution.

Our audit method relies on estimating $\mu$ in $\mu$-GDP and then converting to \adp.
Note that the Gaussian mechanism that adds $\mathcal{N}(0, \sigma^2I)$ noise to the statistic $\theta$, gives $\mu$-GDP if $\sigma = \frac{c}{\mu}$.
This means estimating $\mu$ is equivalent to estimating $\sigma$ given a decision rule and observations from $\mathcal{O}$ and $\mathcal{O}'$. 

For $\mu$-GDP we have 

\begin{align}
\beta_{\alpha} &:= \inf\{\beta_z : \alpha_z \leq \alpha\} \\
&= \Phi(\Phi^{-1}(1-\alpha) - \mu)
\end{align}

where $z$ is a decision threshold on the real line. 
For any $z$ we have:

\begin{align}
\begin{split}
\Phi^{-1}(1-\alpha) - \Phi^{-1}(\beta) &= \Phi^{-1}(1-1+\Phi(\frac{z}{\sigma})) \\&-
\Phi^{-1}(1-\Phi(\frac{z-1}{\sigma})) 
\end{split}\\
&= \frac{z}{\sigma} - \Phi^{-1}(\Phi(\frac{1-z}{\sigma})) \\
&= \frac{1}{\sigma}
\end{align}

That is, given two Gaussian distributions, the lower bound derived from GDP is independent of the threshold. 
Given a sufficient number of observations from $\mathcal{O}$ and $\mathcal{O}'$, our lower bounds found through GDP will be approximately equal, while we must use the threshold $w$ to maximize the lower bound found through \adp auditing.

\begin{figure}[t]
% \captionsetup{width=0.95\columnwidth, justification=centering}
\centering
    \includegraphics[width=1.\linewidth]{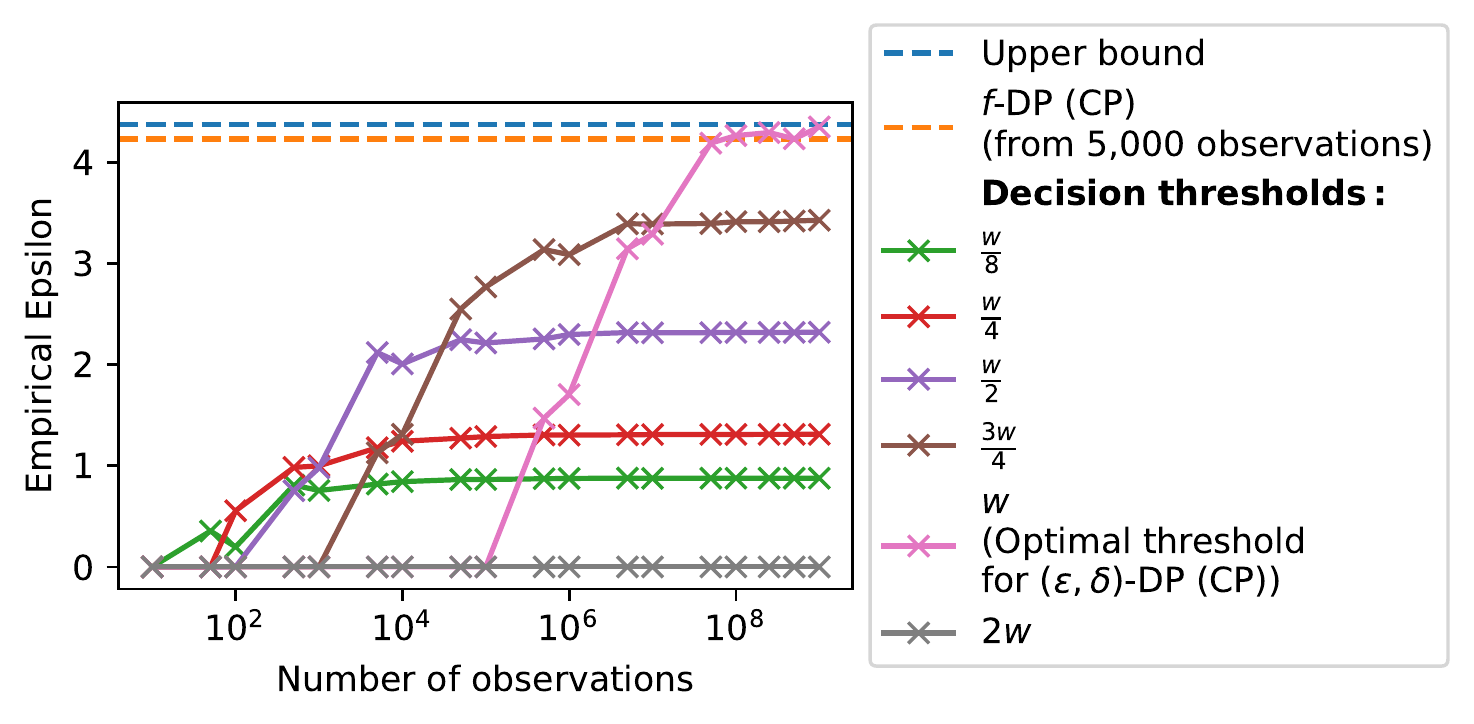}
\caption{We compare how the number of observations affect the lower bound found through ($\epsilon$, $\delta$)-DP with Clopper-Pearson using different decision thresholds. The optimal threshold as set out by \cref{eq:starstar} only reaches the upper bound when the number of observations is >100M. The lower bounds found with other decision thresholds saturate at smaller values than the upper bound.}
\label{fig: threshold_comparison_nsamples_and_eps_delta}
\end{figure}

We empirically show that difference in bounds between our GDP method and ($\epsilon$, $\delta$)-DP with Clopper-Pearson for different thresholds at a sample size of observations equal to $5K$ in \cref{fig: threshold_comparison_nsamples_5000} with $\sigma=1$.
As expected, different decision thresholds lead to approximately the same lower bound found through GDP auditing, while the lower bounds drastically change for ($\epsilon$, $\delta$)-DP auditing with Clopper-Pearson.
Furthermore, the optimal threshold for ($\epsilon$, $\delta$)-DP auditing (given by \cref{eq:starstar}) with Clopper-Pearson gives a lower bound of zero since there aren't a sufficient number of observations to cover this region of the observation space.
% Empirically, 5K samples from $p_0$ and $p_1$ is sufficient for our choice of decision threshold to not matter, and so we use a threshold of $\frac{1}{2}$ in our main experiments.
% We note that the ($\epsilon$, $\delta$)-DP with Clopper-Pearson lower bound varies with different thresholds, as predicted by our theory.
% We also note that the optimal threshold given by \cref{eq:starstar} does not produce a lower bound > 0 for 5K samples.
When auditing with ($\epsilon$, $\delta$)-DP with Clopper-Pearson, we show that the optimal threshold given by \cref{eq:starstar} decision threshold can achieve tight lower bounds given enough samples from $\mathcal{O}$ and $\mathcal{O}'$ in \cref{fig: threshold_comparison_nsamples_and_eps_delta}, while other thresholds cannot, again, as predicted by our theory.

In sum, we can achieve tight lower bounds with GDP, and with as little as 5K samples we can expect any decision threshold to give strong lower bounds, removing our need for hold-out observations to find the optimal threshold.
Using the ($\epsilon$, $\delta$)-DP with Clopper-Pearson lower bound requires us to either (1) use an extremely large number of samples and then use the optimal decision threshold -- this will be prohibitively expensive as this would require > 100M observations, or  (2), use a random / guess threshold which will saturate far below the upper bound for any number of observations.
From \cref{fig: threshold_comparison_nsamples_and_eps_delta}, the maximum lower bound found through Clopper-Pearson requires approximately 250M observations to reach the same lower bound found through auditing with Clopper-Pearson and GDP.
In other words, we would need approximately $5000$x more observations.

\subsection{Choosing a Canary Gradient} \label{ssec:canary_grad_choose}

We first investigate how the canary gradient affects the estimated privacy bound.
We construct the canary gradient in three ways: 
\begin{enumerate}
    \item \text{Dirac canary}: All gradient values are zero except at a single index.
    \item \text{Constant canary}: All gradient values have the same value.
    \item \text{Random canary}: Gradient sampled from a Gaussian.
\end{enumerate}

\begin{figure}[t]
% \captionsetup{width=0.95\columnwidth, justification=centering}
  \centering
\begin{subfigure}{0.45\textwidth}
\centering
    \includegraphics[width=0.95\linewidth]{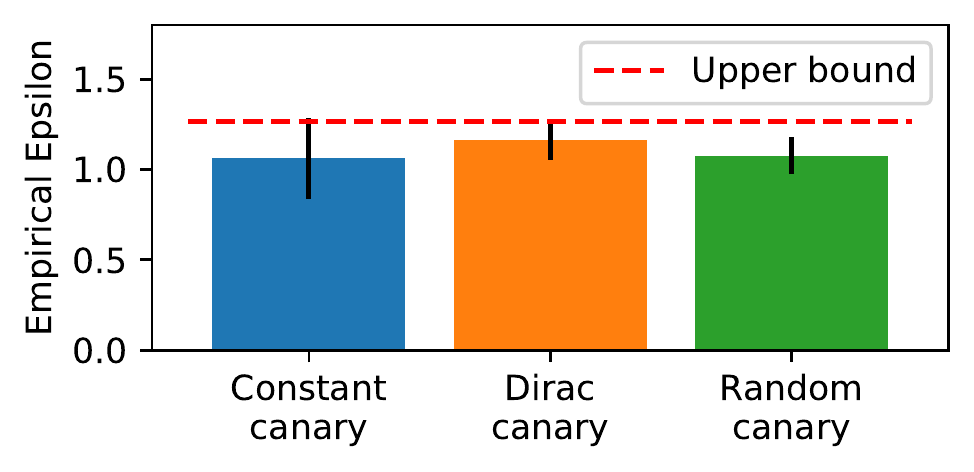}
        \caption{We compare three different ways to create a canary gradient. The Dirac canary gradient, with zeros everywhere except for a single position which has a value set to the clipping norm, slightly outperforms other approaches.}
\end{subfigure}
\begin{subfigure}{0.45\textwidth}
\centering
    \includegraphics[width=0.95\linewidth]{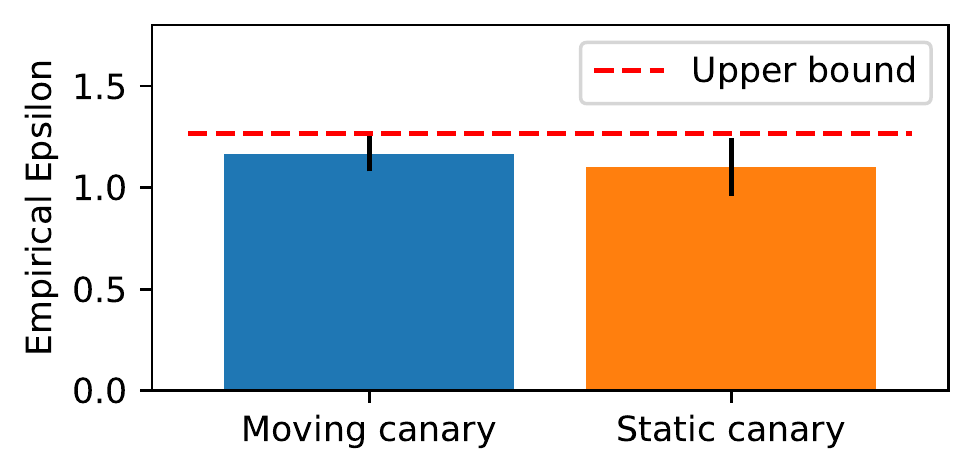}
        \caption{At each iteration we can either insert the same (static) canary gradient, or compute a new canary gradient (moving). There is little difference between these two approaches in terms of the $\epsilon$ lower bound we can find.}
\end{subfigure}%
\caption{How design decisions for the canary gradient change the lower bound we compute for $\epsilon$ with \fdp.}
\label{fig: canary_type}
\end{figure}

In each setting, gradient values are re-scaled such that the canary gradient has a maximum norm equal to the clipping norm.
We also measure if there is a difference between using the same canary gradient at each iteration of DP-SGD or creating a new canary gradient.
For example, in the Dirac canary we would randomly sample a new index to set to the clipping norm at each update.
Results are shown in \cref{fig: canary_type}; using a Dirac canary that is reset at each update performs best.

\subsection{Choosing a Canary Input in The White-box Setting}\label{app:diff_inp}
We evaluate how different types of input canaries can affect auditing in a white-box setting. We construct the canaries in four different ways:
\begin{enumerate}
    \item \textbf{Mislabeled  example}: We select a random example from the test dataset of the model and we select a random label (that is not equal to the original label). 
    \item \textbf{Blank example}: We craft an input where all dimensions of the input are equal to zero. 
    \item \textbf{Adversarial example}: We apply Projected Gradient Descent (PGD) to generate adversarial example on a random example from the test dataset. 
    \item \textbf{Crafted example}: We use~\Cref{alg:wb_canary} to generate an input example.
\end{enumerate}

As we saw from our gradient experiment using the dot product between the privatized gradient and the canary gradient is a sufficient metric for auditing DP-SGD. 
Therefore we use the same idea in our crafting algorithm (\Cref{alg:wb_canary}) in input space. 
We look for a canary such that its gradient is orthogonal to other gradients in the training batch. 
However, we cannot use the example in the training batch directly to craft such an example as it will violate the DP-assumptions (the adversary cannot have access to non-noisy gradients). 
Therefore, we assume the adversary has access to an example from the same distribution as the training dataset and uses that data to estimate the gradient of the model on the training example. 
Then, it crafts an example such that its gradient is orthogonal to the estimated gradient. 
Thus, creating a gradient that is significantly different from other examples in the batch and its presence can be detected. 

\begin{figure}
    \centering
    \includegraphics[scale=0.4]{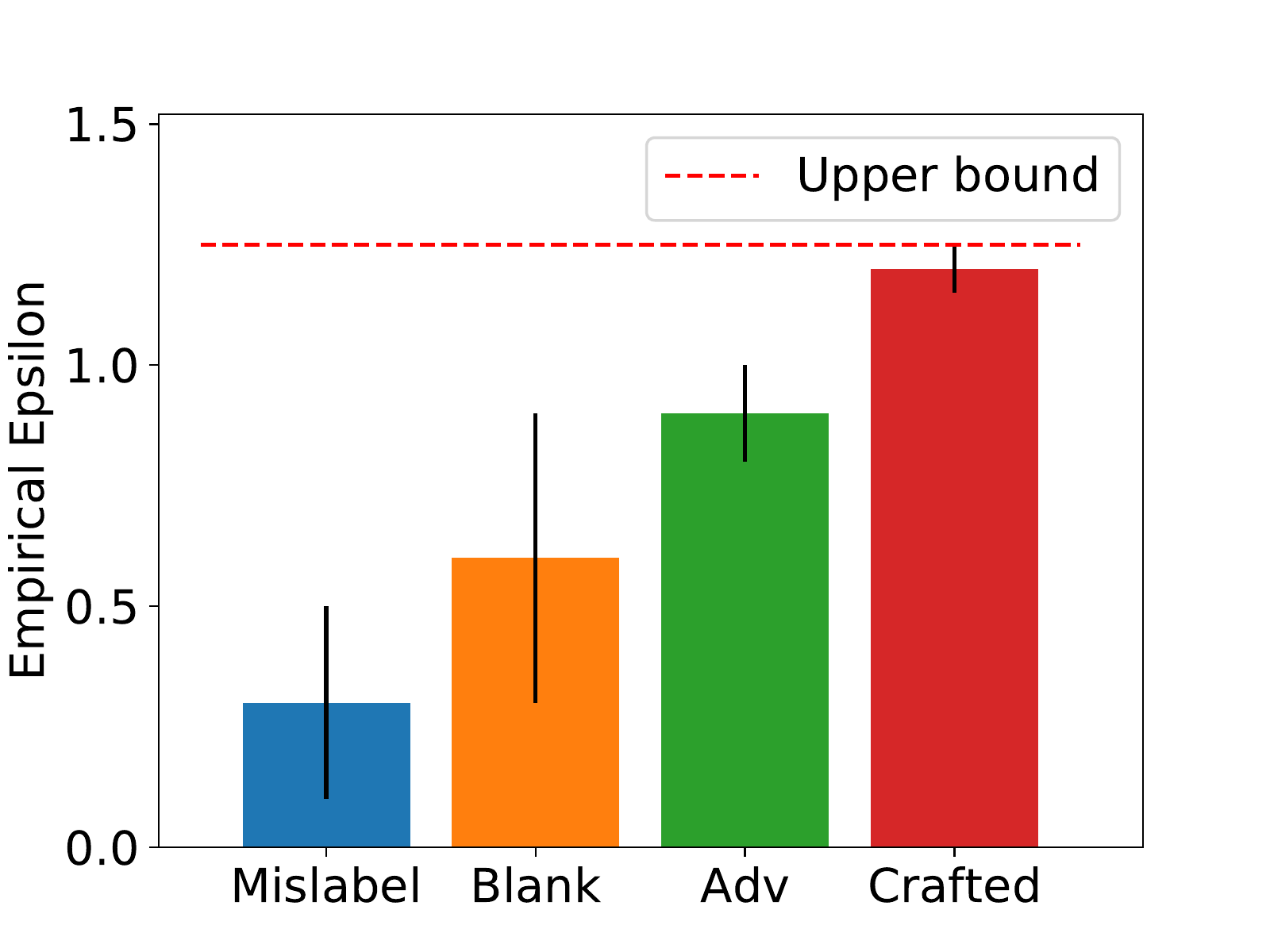}
    \caption{Comparison of the different canary crafting approaches in input space for CIFAR-10 dataset using WRN architecture.}
    \label{fig:diff_input_attacks}
\end{figure}

Figure~\ref{fig:diff_input_attacks} compares the effectiveness of different input canaries in the white-box setting. 
As can be seen, using our canary crafting approach we can achieve significantly tighter bounds on DP-SGD compared to other canary crafting strategies. 
Unfortunately, it is not trivial to extend either the adversarial example or our crafting approach to the black-box setting and therefore we do not use them in the black-box experiments.

% \subsection{Number of Observations}

% The quality of a lower bound is dependent on the number of observations.
% This is another area where auditing with \fdp vastly outperforms \adp.
% % For example, if the attacker has 1000 observations, auditing with \adp with Clopper-Pearson, the maximum lower bound is BLAH, and with \fdp with Clopper-Pearson the maximum lower bound is BLAH.
% In \Cref{fig:audit_gdp_dp} and \cref{app:choose_threshold}, we show that \adp auditing requires between 100-5000x as many observations to give the same lower bound as found with \fdp.
% In general, auditing the Gaussian mechanism with GDP gives tight lower bounds with as little as a few hundred observations.

\section{The effect of model related parameters}
\label{app:parameters}

% TODO
\subsection{Do earlier training steps leak more information?}\label{app:diff_steps}
\begin{figure*}[t]
    \centering
    \begin{subfigure}{0.5\textwidth}
    \centering
    \includegraphics[width=0.8\linewidth]{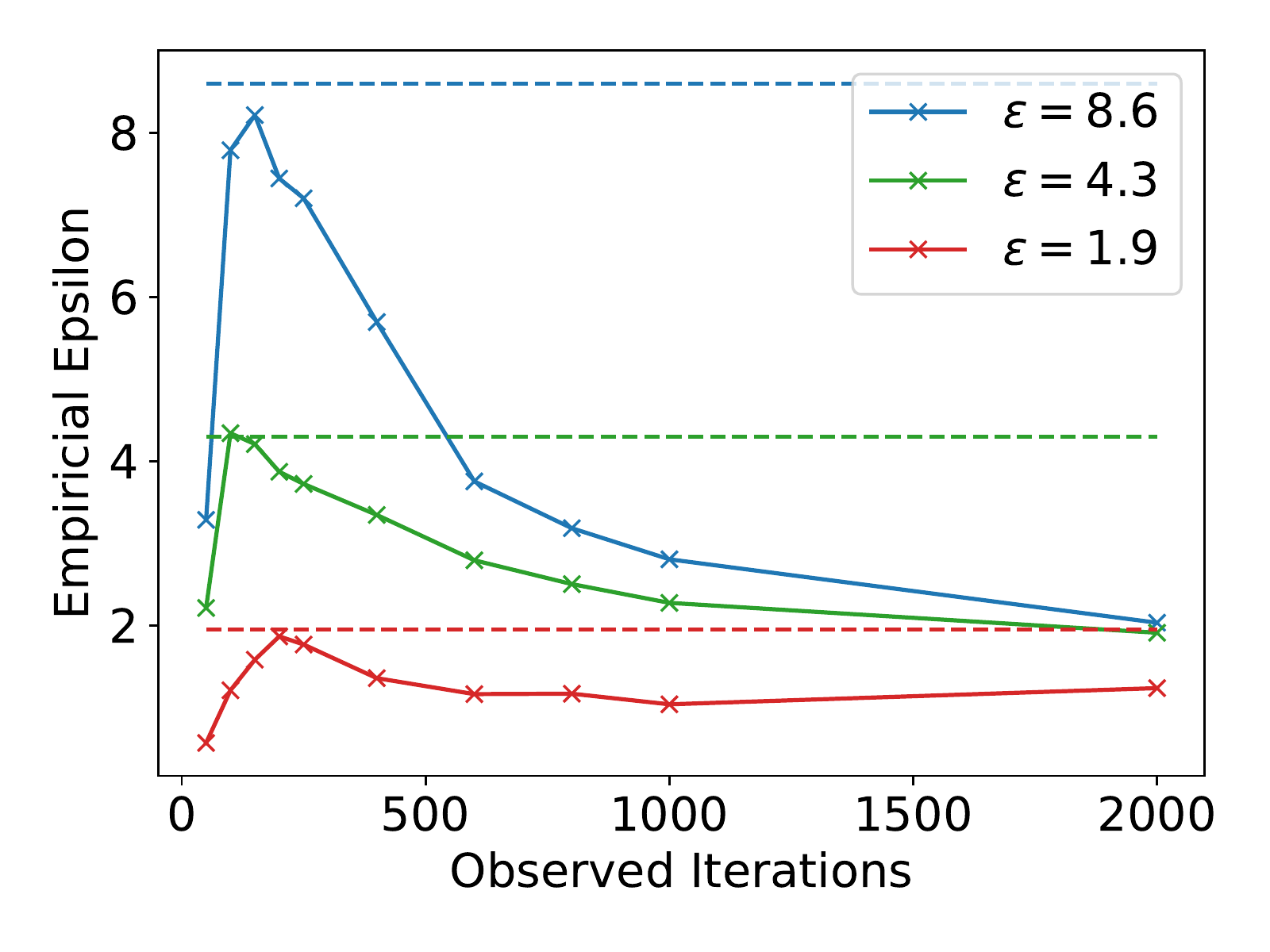}
    \caption{CIFAR-10 dataset with a WRN-16 model.}
    \end{subfigure}
    \begin{subfigure}{0.49\textwidth}
    \centering
    \includegraphics[width=0.8\linewidth]{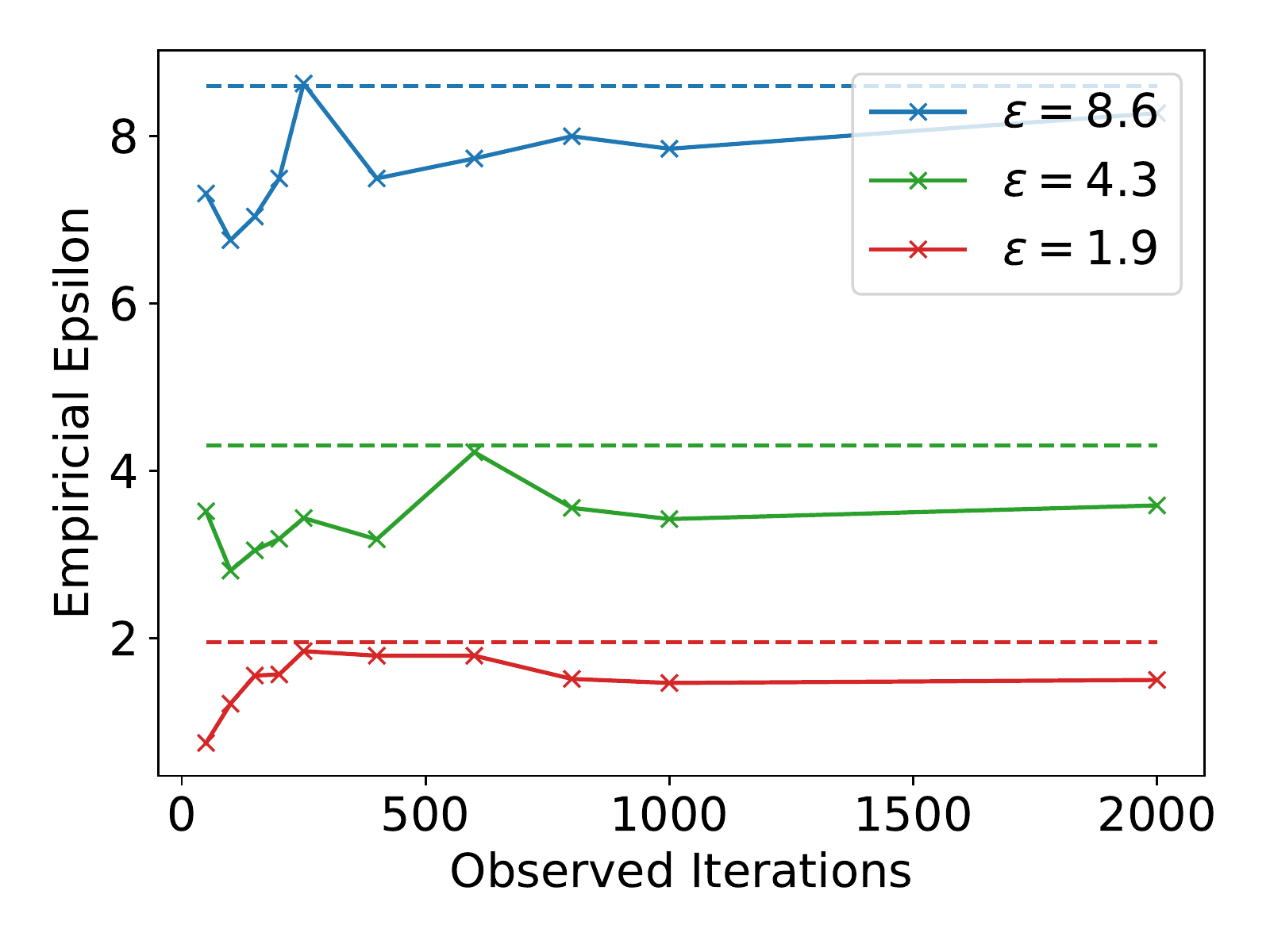}
    \caption{Random dataset with a WRN-16 model.}
    \end{subfigure}
    \caption{Lower bounding $\epsilon$ with the input space attack (white-box setting) for different number of iterations of training and values of $\epsilon$, using \fdp(ZB).}
    \label{fig:audit_diff_steps}
\end{figure*}

Recent work on measuring the amount of privacy leaked when training convex models with DP has shown that the true amount of leakage plateaus as the model converges~\cite{DBLP:journals/corr/abs-2203-05363, DBLP:journals/corr/abs-2205-13710}.
This means that after a certain number of steps, training for more iterations does not consume any more of the privacy budget.
The analysis for this result is specific to convex models; and we cannot prove such properties for the deep models. 
However, when can evaluate if a similar phenomenon holds empirically. 
We measure the lower bounds when we only use the first $n$ iterations of the training in the \emph{White-box access with Input Space Canaries} threat model.
Results are shown in \Cref{fig:audit_diff_steps}.
We find that indeed, the first part of training does leak more information than later in training on the CIFAR-10 dataset. However, when we evaluate the random dataset we do not see the same behavior (please note that, if we only look at a very small number (<100) of the iterations we get a very loose bound on $\epsilon$ because we do not have enough observations to have a sufficiently confident estimation). 
Understanding why we cannot lower bound from our audit becomes looser in later iterations requires further investigation which we leave for future work. 
Nevertheless, the results suggest that when we limit the adversary to canaries in the input space then model architecture, underlying dataset and the how well a model has been trained all have an effect on privacy leakage.

\subsection{Do larger models leak more privacy?}\label{app:diff_sizes}
\begin{figure}[t]
% \captionsetup{width=1.\columnwidth, justification=centering}
\centering
    \includegraphics[width=1.\linewidth]{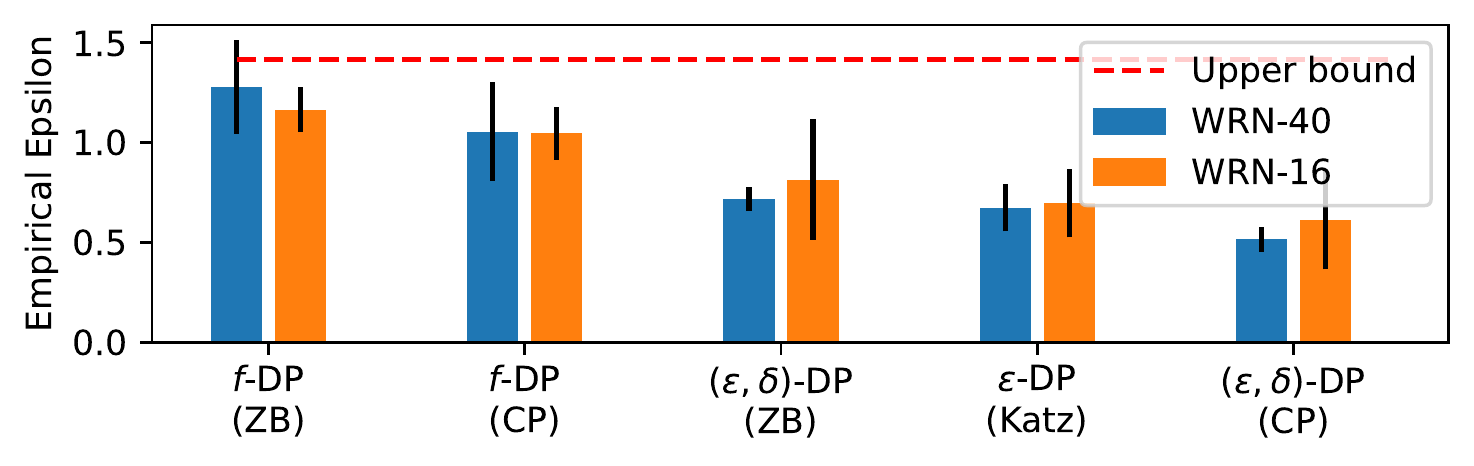}
\caption{Comparison of how model architecture (WRN-16 and -40) affects the $\epsilon$ lower bound.}
\label{fig: wrn_arch}
\end{figure}

% Given that the effect of hyper-parameters on Purchase dataset is not studied even in non-private settings, we focus on CIFAR10 dataset in this section.
Recent work has shown the larger models have a greater capacity to memorize training data verbatim~\cite{carlini2022quantifying}.
We investigate if the same trend holds when training with DP-SGD by comparing lower bounds on a WRN-16 and WRN-40 model.
Results are shown in \Cref{fig: wrn_arch}.
Interestingly, if one was to use a non \fdp auditing method, one would make an incorrect conclusion that the WRN-16 leaks more than the WRN-40.
Using our \fdp auditing method, we identify that, indeed, the larger WRN-40 model leaks slightly more than the WRN-16 model.

\subsection{Does augmentation multiplicity affect privacy lower bounds?}\label{app:augmult}

\begin{figure}[t]
% \captionsetup{width=1.\columnwidth, justification=centering}
\centering
    \includegraphics[width=1.\linewidth]{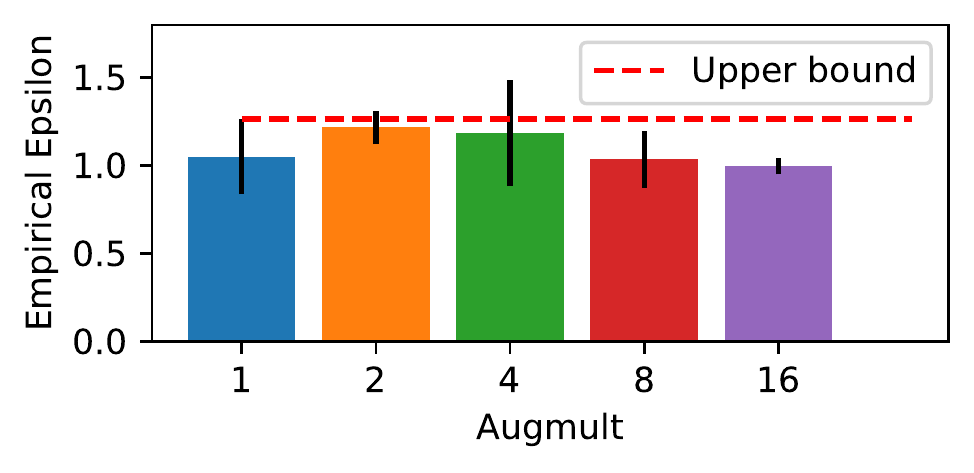}
\caption{How the value for Augmult~\cite{de2022unlocking} affects the $\epsilon$ lower bound.}
\label{fig: augmult}
\end{figure}

In De et al.~\cite{de2022unlocking}, data augmentation was a key ingredient in achieving state-of-the-art results on CIFAR-10.
In particular, De et al. use a data augmentation technique they term \emph{augmentation multiplicity} (Augmult), where they augment a single example multiple times and compute the average gradient over these augmentations \emph{before} clipping. 
They find that increasing the Augmult value (number of augmentations) improves performance; increasing this value does not increase the privacy cost as it does not change the sensitivity of the privatized gradient to any single example in the batch.
In \cref{fig: augmult}, we measure lower bounds at different Augmult values, and observe no clear trend between lower bounds and the value for Augmult.

\end{document}